\documentclass[journal,transmag]{IEEEtran}
\usepackage{multirow}
\usepackage{subfigure}
\usepackage{graphicx}
\usepackage{array}
\usepackage{amssymb}
\usepackage{amsmath}
\usepackage{algorithmic}
\usepackage{color}
\usepackage{cite}

\makeatletter

\let\NAT@parse\undefined

\makeatother

\usepackage{hyperref}  
\usepackage{bm}
\usepackage{amssymb}
\usepackage{latexsym}
\usepackage{dsfont}
\usepackage{multirow}
\usepackage{amsfonts}
\usepackage{pifont}
\usepackage{amsmath}

\newcommand{\tabincell}[2]{\begin{tabular}{@{}#1@{}}#2\end{tabular}}

\def\ie{{\textit{i.e.}}}
\def\eg{{\textit{e.g.}}}

\title{Exploring Optical-Flow-Guided Motion and Detection-Based Appearance for Temporal Sentence Grounding}
\author{Daizong Liu, Xiang Fang, Wei Hu,~\IEEEmembership{Senior~Member,~IEEE}, Pan Zhou,~\IEEEmembership{Senior~Member,~IEEE}
\IEEEcompsocitemizethanks{
\IEEEcompsocthanksitem The first two authors contributed equally to the paper writing and experiments. Corresponding authors: Wei Hu and Pan Zhou.
\IEEEcompsocthanksitem Daizong Liu and Wei Hu are with Wangxuan Institute of Computer Technology, Peking University, No. 128, Zhongguancun North Street, Beijing, China. E-mail: dzliu@stu.pku.edu.cn, forhuwei@pku.edu.cn. 
\IEEEcompsocthanksitem Xiang Fang is with the Hubei Engineering Research Center on Big Data Security, School of Cyber Science and Engineering Huazhong University of Science and Technology, Wuhan 430074, China (e-mail: xfang9508@gmail.com)
\IEEEcompsocthanksitem Pan Zhou  is with the Hubei Engineering Research Center on Big Data Security, School of Cyber Science and Engineering, Huazhong University of Science
and Technology, Wuhan 430074, China (e-mail: panzhou@hust.edu.cn).
}}

\begin{document}
\maketitle
\begin{abstract}
Temporal sentence grounding aims to localize a target segment in an untrimmed video semantically according to a given sentence query. 
Most previous works focus on learning frame-level features of each whole frame\footnote{In this paper, the frame is a general concept for an actual video frame or a video clip which consists of a few consecutive frames.} in the entire video, and directly match them with the textual information.
Such frame-level feature extraction leads to the obstacles of these methods in distinguishing ambiguous video frames with complicated contents and subtle appearance differences, thus limiting their performance.
In order to differentiate fine-grained appearance similarities among consecutive frames, some state-of-the-art methods additionally employ a detection model like Faster R-CNN to obtain detailed object-level features in each frame for filtering out the redundant background contents.
However, these methods suffer from missing motion analysis since the object detection module in Faster R-CNN lacks temporal modeling.
To alleviate the above limitations, in this paper, we propose a novel \textbf{M}otion- and \textbf{A}ppearance-guided \textbf{3}D \textbf{S}emantic \textbf{R}easoning \textbf{N}etwork (MA3SRN), which incorporates optical-flow-guided motion-aware, detection-based appearance-aware, and 3D-aware object-level features to better reason the spatial-temporal object relations for accurately modelling the activity among consecutive frames.
Specifically, we first develop three individual branches for motion, appearance, and 3D encoding separately to learn fine-grained motion-guided, appearance-guided, and 3D-aware object features, respectively. 
Then, both motion and appearance information from corresponding branches are associated to enhance the 3D-aware features for the final precise grounding. 
Extensive experiments on three challenging datasets (ActivityNet Caption, Charades-STA and TACoS) demonstrate that the proposed MA3SRN model achieves a new state-of-the-art.  
\end{abstract}

\section{Introduction}
As an important multimedia task of cross-modal understanding, temporal sentence grounding (TSG) aims to retrieve the most relevant video segment according to a given sentence query \cite{wang2021weakly,tang2021frame,liu2022exploring,liu2022memory,liu2022unsupervised}. 
There are several related tasks proposed involving both video and language, such as 
temporal action localization
\cite{shou2016temporal,zhao2017temporal,sun2021exploiting}, 
video question answering \cite{gao2019structured,le2020hierarchical,wang2021dualvgr},
and video captioning \cite{jiang2018recurrent,chen2020learning,xu2020deep}. 
Compared with these tasks, TSG is more challenging as it needs to not only capture the complicated visual and textual information, but also learn the complex multi-modal interactions between visual and textual information for accurately modelling the target activity.

\begin{figure}[t]
\centering
\includegraphics[width=0.48\textwidth]{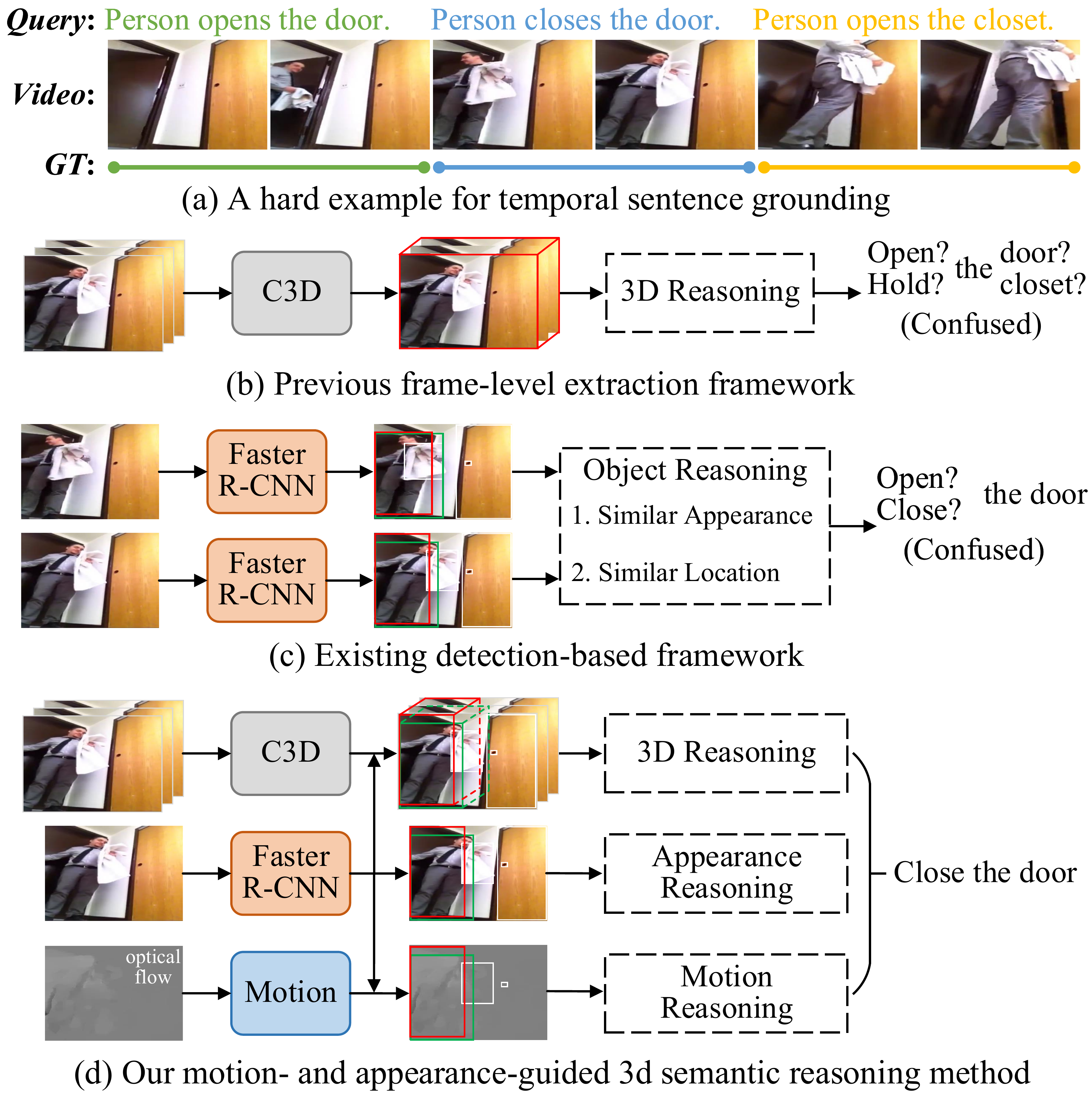}
\caption{(a) A hard example of temporal sentence grounding where the video contains several semantically similar segments. (b) Previous framework typically extracts frame-level features for reasoning, which fails to capture subtle appearance differences (``door" and ``closet") and distinguish the complex foreground-background contexts (``open" and ``hold") in each frame. (c) Existing detection-based framework only extracts appearance-aware object information, and fails to distinguish similar motions ``open" and ``close". (d) Our method develops three specific branches to separately learn optical-flow-guided motion, detection-based appearance and 3D-aware object contexts, and associates the information of them to better reason the query.}
\label{fig:intro}
\end{figure}

Most existing works \cite{anne2017localizing,gao2017tall,chen2018temporally,zhang2019cross,liu2020reasoning,liu2020jointly,zhang2019learning,liu2021adaptive,liu2021progressively} develop a proposal-ranking framework that first generates multiple candidate segment proposals and then ranks them based on their similarities with the sentence query. The segment proposal with the highest similarity score is finally selected as the best target segment. 
Instead of using complex proposals, some recent works \cite{rodriguez2020proposal,chenrethinking,yuan2019find,mun2020local,zhang2020span} utilize a proposal-free approach that directly regresses the temporal locations of the target segment. Specifically, they either regress the start/end timestamps based on the entire video representation, or predict at each frame to determine whether this frame is the start or end boundary.
Compared to the proposal-ranking counterparts, these proposal-free methods have faster running speed but generally achieve lower performance. 
Although the above two types of methods have achieved impressive results, all of them attempt to extract the frame-level visual features for each {\it whole} frame to model the semantic of the target activity, which captures the redundant background information and fails to explore the fine-grained differences among video frames with high similarity. This failure is significantly serious in adjacent frames near the segment boundary. 
For instance, for two queries with similar semantics (\eg, ``Person opens the door" and ``Person opens the closet" as shown in Fig. \ref{fig:intro}(a) and (b)), modeling the temporal relations by frame-level features can capture the same action ``open", but it is difficult to adequately distinguish the subtle details of different objects (``door" and ``closet"). Besides, it also fails to distinguish the complex foreground (``open" the door) and background (``hold" the clothes) activity contents with the same appearing object ``person".

Recently, in order to capture more fine-grained foreground object-level appearance features in each frame, some detection-based methods \cite{zeng2021multi,zhang2020object,zhang2020does} have been proposed and achieved promising results. 
To accurately reason the semantic of the target activity for modelling, \cite{zeng2021multi} considers temporal sentence grounding and learns spatio-temporal object relations, while other methods \cite{zhang2020object,zhang2020does} focus on the spatio-temporal object grounding task which aims to retrieve an object rather than a video segment.
By learning the object-level representations in each frame, these methods significantly alleviate the issue of indistinguishable local appearances and foreground-background contents, such as ``door" and ``closet", ``open" and ``hold". 
However, methods like \cite{zeng2021multi} generally extract object features by an object detection model (\ie, Faster R-CNN \cite{ren2015faster}), which lacks the object-level motion context for the temporal action modelling of a specific object (\eg, ``door" and ``closet"), thus degenerating the performance on semantically similar events. 
For example, as shown in Fig. \ref{fig:intro}(c), 
although detection-based methods can learn the object appearance in each frame, 
they have difficulties in distinguishing the similar motions ``open" and ``close" by learning the object relations in consecutive frames, since the detected objects extracted by Faster R-CNN have similar appearance and spatial positions in these frames. 
Since the motion context is necessary to model the consecutive states or actions for objects \cite{liu2022investigating}, how to effectively encode and integrate the motion context and the appearance information to compose the complicated activity is a crucial problem in TSG.

To this end, we propose a novel \textbf{M}otion- and \textbf{A}ppearance-guided \textbf{3}D \textbf{S}emantic \textbf{R}easoning \textbf{N}etwork (MA3SRN),
which incorporates both motion contexts and appearance contexts into 3D-aware object features for better modelling the target activity.
Considering the optical flow is widely used in various video understanding tasks \cite{kangaspunta2021adaptive,blattmann2021understanding,dorkenwald2021stochastic,woo2021learning,jayasundara2021flowcaps} which identifies actions with large motion, we extract optical flows between adjacent frames among the entire video offline and take them as the motion information. 
To efficiently capture the appearance information, we obtain the object-level appearance from a detection model (\eg, Faster R-CNN).
Also, we follow previous TSG works to encode 3D-aware features from the video clips with a C3D network \cite{tran2015learning} to embed the spatio-temporal information. 
Particularly, we apply the bounding boxes (bbox) extracted from the detection model on the motion and 3D-aware features to capture the subtle object information for filtering out the redundant backgrounds. 
After obtaining the above three kinds of features, we integrate their contexts to generate more representative features.

However, it is ineffective to directly construct a multi-stream model that takes individual features (\ie, motion, appearance, or 3D-aware feature) as the input in each stream and subsequently concatenate them as the multi-stream output, because this lacks the interaction among these different features. 
Considering that motion, appearance, and 3D-aware features are complementary to each other, we design an effective and novel fusion scheme to dynamically integrate different kinds of features to enhance the overall feature representations for improving the learning of each modality.

Specifically, we first detect and obtain appearance-aware object representations by a Faster R-CNN model, and simultaneously employ the RoIAlign \cite{he2017mask} on the optical-flow-guided and 3D-aware feature maps from the ResNet50 \cite{he2016deep} and C3D network for the extraction of motion-aware and 3D-aware object features, respectively. 
Then, we develop three separate branches to reason the motion-guided, appearance-guided, and 3D-aware object relations, respectively. 
In each branch, we interact object features with query information for query-related object semantic learning, and adopt a fully-connected object graph for spatio-temporal semantic reasoning. 
At last, we represent frame-level features by aggregating object features inside the frame, and introduce an attention-based associating module to selectively integrate representative information from three branches for the final grounding.

This paper extends our previous work \cite{liu2022exploring} in three aspects. 
Firstly, the motion information in \cite{liu2022exploring} captures only partial motion, since it is directly extracted from the C3D network that contains both spatial-aware appearance context and temporal-aware action context. In contrast, this paper takes such embeddings from the C3D network as the \textbf{3D-aware features} for spatial-temporal context encoding, and represents the {\it exact} motion information via optical flows, with a reasoning branch integrated for action inferring. 
Secondly, we design a more powerful motion-appearance associating module with co-attentional transformers \cite{lu2019vilbert} to associate and incorporate the motion-guided, appearance-guided features into the 3D-aware one for spatial-temporal object-level representation learning. 
Thirdly, we add experiments on a large-scale dataset ActivityNet Caption \cite{krishna2017dense}, and conduct more experiments to demonstrate the effectiveness of the proposed MA3SRN.

The contributions of this paper are summarized as follows:
\begin{itemize}
    \item As far as we know, we are the first to propose a Motion- and Appearance-guided 3D Semantic Reasoning Network for Temporal Sentence Grounding, exploring the optical-flow-guided motion-aware,  detection-based appearance-aware, and 3D-aware object features.
    \item We devise motion, appearance, and 3D-aware branches separately to capture action-oriented, appearance-guided, and 3D-aware object relations. An attention-based motion-appearance associating module is further proposed to integrate the most representative features from these three branches for the final grounding.
    \item We conduct extensive experiments on three challenging datasets: ActivityNet Caption, Charades-STA and TACoS. Experimental results show that our MA3SRN outperforms state-of-the-art approaches by a large margin.
\end{itemize}

\section{Related Work}
\subsection{Image/Video Retrieval}
Given a set of candidate images/videos and a sentence query, early retrieval works mainly aim to select the image/video that best matches the query \cite{bai2021unsupervised,chen2021feature,weng2021online,song2021spatial,qi2021semantics}. As for image retrieval, previous works focus on localizing the semantically relevant image region. They first generate candidate image regions using image proposal method \cite{ren2015faster}, and then find the matched one with respect to the given query. Some works \cite{mao2016generation,hu2016natural,rohrbach2016grounding} try to extract target image regions based on description reconstruction error or probabilities. 
As for video retrieval,  some methods \cite{otani2016learning,xu2015jointly} incorporate deep video-language embeddings to boost retrieval performance, similar to the image-language embedding approach \cite{socher2014grounded}. Lin \textit{et al.} \cite{lin2014visual} first parse the query descriptions into a semantic graph and then match them to visual concepts in videos. 

\subsection{Temporal Action Localization}
As a single-modal task, temporal action localization aims to classify action instances by predicting the corresponding start timestamps, end timestamps, and action category labels \cite{sun2021exploiting,pramono2021spatial,zhai2021action}. Existing methods can be divided into one-stage methods \cite{lin2017single, long2019gaussian, xu2020g} and two-stage methods \cite{xu2017r, chao2018rethinking, lin2019bmn,xu2019two}. The one-stage methods directly predict action boundaries and labels simultaneously. Specially, Xu \textit{et al.} \cite{xu2020g} employ a graph convolutional network to perform one-stage action localization. In contrast, the two-stage methods first generate action proposals, then refine and classify confident proposals. Generally, 
these confident proposals are generated by the anchor mechanism \cite{xu2017r, chao2018rethinking,xu2019two, yang2020revisiting}. Besides the anchor-based mechanism, we can generate proposals by other technologies, such as sliding window \cite{shou2016temporal}, temporal actionness grouping \cite{zhao2017temporal}, combining confident starting and ending frames \cite{lin2018bsn, lin2019bmn}. 

\subsection{Temporal Sentence Grounding}
As an important yet challenging multimedia task introduced by \cite{gao2017tall} and \cite{anne2017localizing}, temporal sentence grounding (TSG) tries to identify the boundary of the specific video segment semantically corresponding to a given sentence query. Different from temporal action localization, TSG is substantially more challenging as it needs not only capture the complicated visual and textual information, but also learn the complex multi-modal interactions among them for modelling the target activity. Traditional methods \cite{liu2018attentive,gao2017tall} localize the target segment via generating video segment proposals. They sample candidate segments from a video first, and subsequently integrate query with segment representations via a matrix operation. 
These methods lack a comprehensively structure for effective multi-modal features interaction. 
To further mine the cross-modal interaction more effectively, some works \cite{xu2019multilevel,chen2019semantic,ge2019mac,zhang2019learninga} integrate the sentence representation with those video segments individually, and then evaluated their matching relationships.
For instance, Xu \textit{et al.} \cite{xu2019multilevel} introduce a multi-level model to integrate visual and textual features earlier and further re-generate queries as an auxiliary task.
Chen \textit{et al.} \cite{chen2018temporally} capture the evolving fine-grained frame-by-word interactions between video and query to enhance the video representation understanding.
Other works \cite{wang2019temporally,zhang2019cross,zhang2019man,yuan2019semantic,liu2022memory,wang2021weakly} propose to directly integrate sentence information with fine-grained video clip, and predict the temporal boundary of the target segment by gradually merging the fusion feature sequence over time. 

Different from these above methods based on segment candidates, some works \cite{rodriguez2020proposal,chenrethinking,yuan2019find,mun2020local,zhang2020span,liu2022unsupervised,tang2021frame} directly leverage the interaction between video and sentence to predict the starting and ending frames. These works first interact video clips and sentence query information to generate contextual and fine-grained video representation, and then either regress the start/end timestamps based on the video representation or predict at each frame to determine whether this frame is a start or end boundary. 
There are also some reinforcement learning (RL) based frameworks \cite{hahn2019tripping,wu2020tree} proposed in the TSG task.

\begin{figure*}[t]
\centering
\includegraphics[width=0.9\textwidth]{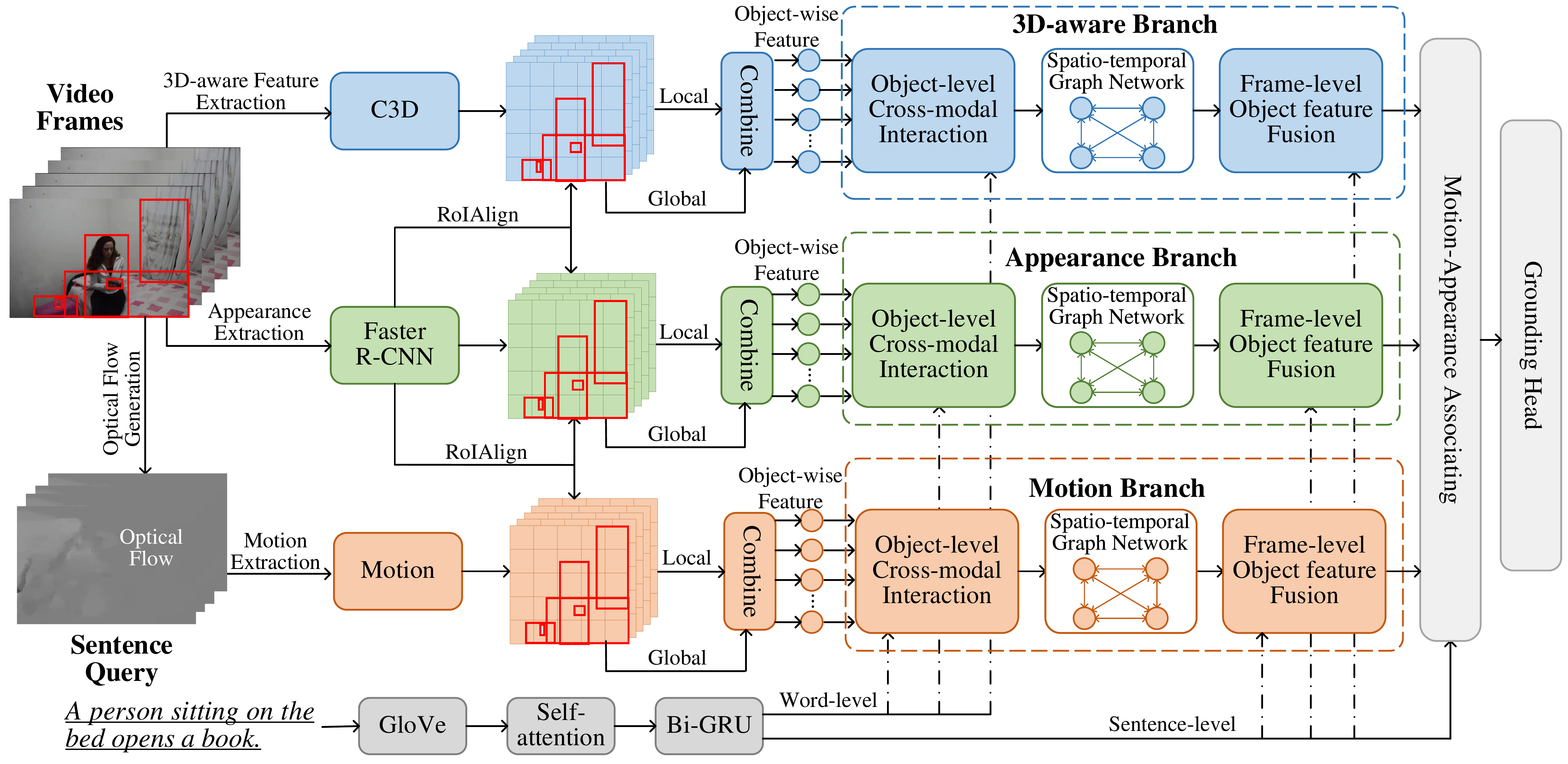}
\caption{The overall pipeline of our proposed MA3SRN model. Firstly, we leverage three-stream video encoders to extract optical-flow-guided  motion,  detection-based  appearance, and 3D-aware object features, respectively. We also utilize the query encoder to embed the word- and sentence-level query features. Then, we develop separate motion, appearance, and 3D-aware  branches for specific cross-modal object reasoning. Finally, we associate motion-appearance contexts and incorporate them into 3D-aware information to generate more representative features for better grounding.}
\label{fig:pipeline}
\end{figure*}

However, the above two types of methods are all based on frame-level features to capture the semantic of video activities, which is unable to capture the fine-grained discrepancy among different frames with high similarity, especially the adjacent frames near the segment boundary.
Recently, some detection-based methods \cite{zeng2021multi,zhang2020object,zhang2020does,liu2022exploring} have been proposed to capture subtle and fine-grained object appearances in each frame, which filters out the background contents and improves the localization performance. However, these methods only extract object features by detection models, thus failing to learn the motion information of each object.



\section{Our Approach}
In this section, we first introduce the problem statement of temporal sentence grounding (TSG). Next, we present the modules of our proposed MA3SRN model, including the video and query encoders, three-stream reasoning branches (3D-aware branch, appearance branch, and motion branch), motion-appearance associating module, and grounding head, as shown in Fig.~\ref{fig:pipeline}. 

\subsection{Problem Statement}
In the TSG task, we are provided with an untrimmed video $\mathcal{V}=\{v_t\}_{t=1}^T$ and a sentence query $\mathcal{Q}=\{q_n\}_{n=1}^N$, where $v_t$ denotes the $t$-th frame, $T$ denotes the frame number, $q_n$ denotes the $n$-th word and $N$ denotes the word number. This task aims to localize the precise start and end timestamps of a specific segment in the video $\mathcal{V}$, which refers to the corresponding semantic of the query $\mathcal{Q}$.

\subsection{Overview}
In this section, we propose a novel  Motion-  and  Appearance-guided 3D Semantic Reasoning Network (MA3SRN), which incorporates optical-flow-guided  motion-aware, detection-based appearance-aware, and 3D-aware object features to better reason the object relations for accurately modelling the activity among consecutive frames. 
As shown in Fig. \ref{fig:pipeline}, we first design three video encoders to extract corresponding optical-flow-guided motion, detection-based appearance, and more contextual spatial-temporal 3D-aware object features, respectively.
Then, we develop separate motion, appearance and 3D-aware branches to learn fine-grained
action-oriented, appearance-aspect and  3D-aware object relations. In each specific branch, after interacting object-query features to filter out irrelevant object features, we reason the relations between the foremost objects with a spatio-temporal graph and represent frame-level features by fusing its contained object features. 
Finally, we associate the frame-level features from three branches, and integrate the motion-appearance information into 3D-aware ones via triple-modal transformers to generate more representative features for the final accurate grounding.

\subsection{Video and Query Encoders}
\subsubsection{Video encoder}
Most previous works employ a general C3D network \cite{tran2015learning} to extract 3D-aware features from each whole frame, which fails to distinguish the visually similar background contents and capture the subtle details of small objects. Although detection-based models utilize Faster R-CNN \cite{ren2015faster} to extract appearance-aware object features on a pre-trained image detection dataset, they lack temporal modeling. Different from these models, we generate both 3D-aware and appearance features to encode more contextual information, and attempt to additionally extract optical-flow-guided motion-aware information to obtain action-oriented features for temporal modeling. We also apply the bounding boxes detected from Faster R-CNN on the above three kinds of features to extract fine-grained object features for filtering out the complicated backgrounds, leading to better multi-modal representation learning.

\noindent \textbf{Appearance feature encoding.} 
For appearance features, we first sample fixed $T$ frames from the original untrimmed video uniformly. Then, based on the Faster R-CNN model built on a ResNet50 \cite{he2016deep} backbone, we obtain $K$ objects from each frame.
Thus, we obtain $T \times K$ objects in total in a single video, and represent their appearance features as $\bm{V}^a_{local}= \{\bm{o}^a_{t,k}, \bm{b}_{t,k}\}_{t=1,k=1}^{t=T,k=K}$, where $\bm{o}^a_{t,k} \in \mathbb{R}^{D}$ denotes the local object-level appearance feature of the $k$-th object in $t$-th frame, $D$ is the feature dimension, and $\bm{b}_{t,k} \in \mathbb{R}^{4}$ represents the corresponding bounding-box position. 
Considering that the global feature of the whole frame also contains the non-local information of its internal objects, we employ another ResNet50 model with a linear layer to generate the global frame-level appearance representation $\bm{V}_{global}^a \in \mathbb{R}^{T \times D}$.

\noindent\textbf{Motion feature encoding.} For motion features, we first generate the optical flows among the frames in the entire video by an energy minimisation framework \cite{simonyan2014two},
and then extract the feature maps of each optical flow through a ResNet50 \cite{he2016deep} model. We apply RoIAlign \cite{he2017mask} on such feature maps and use object bounding-box locations $\bm{b}_{t,k}$ to generate motion-aware object features as $\bm{V}^m_{local}= \{\bm{o}^m_{t,k}, \bm{b}_{t,k}\}_{t=1,k=1}^{t=T,k=K}$. To extract the global features $\bm{V}_{global}^m \in \mathbb{R}^{T \times D}$ of each optical flow, we directly apply average pooling and linear projection to the extracted feature maps of the ResNet50 model.

\noindent\textbf{3D-aware feature encoding.} For 3D-aware features, we first extract the feature maps of each video clip using the last convolutional layer in the C3D \cite{tran2015learning} network. Then, we adopt RoIAlign \cite{he2017mask} on these feature maps and use object bounding-box locations $\bm{b}_{t,k}$ to generate 3D-aware object features $\bm{V}^{3d}_{local}= \{\bm{o}^{3d}_{t,k}, \bm{b}_{t,k}\}_{t=1,k=1}^{t=T,k=K}$. To extract the clip-level global features $\bm{V}_{global}^{3d} \in \mathbb{R}^{T \times D}$, we directly apply average pooling and linear projection to the extracted feature maps of C3D.

For our TSG task,
it is necessary to consider both spatial and temporal locations of each object to reason object-wise relations for accurately modelling the target activity. Therefore, we add a spatio-temporal position encoding to object-level local features in the above three representations as:
\begin{align}
    &\bm{v}_{t,k}^a = \text{FC}([\bm{o}^a_{t,k};\bm{e}^b;\bm{e}^t]), \nonumber\\
    &\bm{v}_{t,k}^m = \text{FC}([\bm{o}^m_{t,k};\bm{e}^b;\bm{e}^t]),\\ 
    &\bm{v}_{t,k}^{3d} = \text{FC}([\bm{o}^{3d}_{t,k};\bm{e}^b;\bm{e}^t]),\nonumber
\end{align}
where $\bm{e}^b = \text{FC}(\bm{b}_{t,k})$ is utilized to encode the spatial position knowledge of each bounding-box, $\text{FC}(\cdot)$ is the fully connected layer, and $\bm{e}^t$ is the temporal position knowledge obtained by \cite{mun2020local} based on the index of each frame. Therefore, the position-aware local object features are denoted as $\widehat{\bm{V}}_{local}^a=\{\bm{v}_{t,k}^a\}_{t=1,k=1}^{t=T,k=K}, \widehat{\bm{V}}_{local}^m=\{\bm{v}_{t,k}^m\}_{t=1,k=1}^{t=T,k=K}, \widehat{\bm{V}}_{local}^{3d}=\{\bm{v}_{t,k}^{3d}\}_{t=1,k=1}^{t=T,k=K}$. Similarly, we add the temporal position encoding into three global representations as:
\begin{align}
   &\widehat{\bm{V}}^a_{global} = \text{FC}([\bm{V}^a_{global};\textbf{e}^T]),\nonumber\\
   & \widehat{\bm{V}}^m_{global} =\text{FC}([\bm{V}^m_{global};\textbf{e}^T]),\\
    &\widehat{\bm{V}}^{3d}_{global} =\text{FC}([\bm{V}^{3d}_{global};\textbf{e}^T]).\nonumber
\end{align}
Finally, we expand the dimension of the above three global features from $T \times D$ to $(T \times K) \times D$, and concatenate the local object-level features with corresponding global features to reflect the context in objects as:
\begin{align}
    &\bm{F}^a = \text{FC}([\widehat{\bm{V}}_{local}^a;\widehat{\bm{V}}^a_{global}]),\nonumber\\
   & \bm{F}^m = \text{FC}([\widehat{\bm{V}}_{local}^m;\widehat{\bm{V}}^m_{global}]),\\
    &\bm{F}^{3d} = \text{FC}([\widehat{\bm{V}}_{local}^{3d};\widehat{\bm{V}}^{3d}_{global}]),\nonumber
\end{align}
where $\bm{F}^a=\{\bm{f}_{t,k}^a\}_{t=1,k=1}^{t=T,k=K} \in \mathbb{R}^{(T \times K)\times D},\bm{F}^m = \{\bm{f}_{t,k}^m\}_{t=1,k=1}^{t=T,k=K} \in \mathbb{R}^{(T \times K)\times D},\bm{F}^{3d} = \{\bm{f}_{t,k}^{3d}\}_{t=1,k=1}^{t=T,k=K} \in \mathbb{R}^{(T \times K)\times D}$ denote the final encoded object-level features.

\subsubsection{Query encoder}
Following previous works \cite{gao2017tall,liu2020jointly,liu2021context}, we first employ the Glove model \cite{pennington2014glove} to embed each word of the given sentence query into a dense vector. 
Then, we use multi-head self-attention \cite{vaswani2017attention} and Bi-GRU \cite{chung2014empirical} modules to encode its sequential information. We denote the final word-level features as $\bm{Q}=\{\bm{q}_n\}_{n=1}^N \in \mathbb{R}^{N \times D}$.
By concatenating the outputs of the last hidden unit in Bi-GRU with a further linear projection, we can obtain the sentence-level feature as $\bm{q}_{global} \in \mathbb{R}^{D}$.

\subsection{Cross-modal Object Reasoning}
After extracting the appearance-aware, motion-aware and 3D-aware object representations, it is necessary to capture the relations between appearance-aware objects for inferring the spatial semantic, learn the relations between motion-aware objects for modeling the temporal semantic. Therefore, 
we develop three separate branches to reason the appearance-guided, motion-guided,  and 3D-aware  object relations with cross-modal interaction. Specifically, in each branch, we first interact object features with the query to enhance their semantic for distinguishing the query-relevant and query-irrelevant objects, and then reason the foremost object relations in spatial temporal spaces within a spatio-temporal graph. A query-guided attention module is further developed to fuse the object information within each frame to represent the frame-level features of current branch.

\noindent \textbf{Cross-modal interaction.}
Learning correlations between visual features and query information is important for query-based video grounding, which helps to highlight the relevant object features corresponding to the query while weakening the irrelevant ones. In details, for the $k$-th object in the $t$-th frame in the appearance branch, we employ a multi-modal interaction module that selectively injects textual evidences into its features. We first utilize an attention mechanism to aggregate the word-level query features $\{\bm{q}_n\}_{n=1}^N$ for each object $\bm{f}^a_{t,k}$ as:
\begin{equation}
    \bm{M}^a_{t,k,n} = \bm{\text{w}}^{\top} \text{tanh}(\bm{W}_1^a \bm{f}^a_{t,k} + \bm{W}_2^a \bm{q}_n + \bm{b}^a_1),
\end{equation}
\begin{equation}
    (\bm{f}^a_{t,k})' = \sum_{n=1}^N \text{softmax}(\bm{M}^a_{t,k,n}) \bm{q}_n,
\end{equation}
where $\bm{W}_1^a$ and $\bm{W}_2^a$ are projection matrices, $\bm{b}^a_1$ is the bias vector and the $\bm{\text{w}}^{\top}$ is the row vector as in \cite{zhang2019cross}. $(\bm{f}^a_{t,k})'$ is the object-aware textual features for the $k$-th object in $t$-th clip. Next, we build the textual gate that takes such textual information as the guidance to weaken the query-irrelevant objects while highlight the query-relevant ones. The query-enhanced object features $\widehat{\bm{f}}^a_{t,k}$ can be obtained by:
\begin{equation}
    \widehat{\bm{f}}_{t,k}^a= \sigma(\bm{W}_3^a (\bm{f}^a_{t,k})'+\bm{b}^a_2) \odot \bm{f}^a_{t,k},
\end{equation}
where $\sigma$ is the sigmoid function, $\odot$ represents element-wise product, $\bm{W}_3^a,\bm{b}_2^a$ are learnable parameters. $\widehat{\bm{F}}^a=\{\widehat{\bm{f}}_{t,k}^a\}_{t=1,k=1}^{t=T,k=K} \in \mathbb{R}^{(T \times K)\times D}$, and the enhanced object features $\widehat{\bm{F}}^m$ and $\widehat{\bm{F}}^{3d}$ of motion and 3D-aware branches can be obtained in the same way.

\noindent \textbf{Spatio-temporal graph reasoning.}
Considering that the query-relevant objects have both spatial interactivity and temporal continuity within the video, as shown in Fig. \ref{fig:pipeline}, we construct a reasoning graph network over all objects to capture their spatio-temporal relations in each branch, respectively. Specifically, for appearance branch, we define object-wise features $\widehat{\bm{F}}^a=\{\widehat{\bm{f}}_{t,k}^a\}_{t=1,k=1}^{t=T,k=K}$ including all objects in all frames as nodes and build a fully-connected appearance graph on them. We adopt graph convolution network (GCN) \cite{kipf2016semi} with leanrable adjacent matrix to learn the spatio-temporal object-relation features via message propagation.
In details, we first measure the pairwise affinity between all object features by:
\begin{equation}
    \bm{A}^a = \text{softmax}((\widehat{\bm{F}}^a \bm{W}^a_4)(\widehat{\bm{F}}^a \bm{W}^a_5)^{\top}),
\end{equation}
where $\bm{W}^m_4,\bm{W}^m_5$ are learnable parameters. $\bm{A}^m \in \mathbb{R}^{(T \times K) \times (T \times K)}$ is the adjacent matrix obtained by calculating the affinity edge of each pair of objects. Two objects with strong semantic relationships will be highly correlated and have an edge with high affinity score in $\bm{A}^a$.
In contrast, two objects with weak semantic relationships will be weakly correlated and have an edge with low affinity score in $\bm{A}^a$.
After obtaining $\bm{A}^a$, we apply one single-layer GCN with residual connections to perform spatio-temporal object semantic reasoning as:
\begin{equation}
    \widetilde{\bm{F}}^a = (\bm{A}^a \widehat{\bm{F}}^a \bm{W}^a_6) \bm{W}^a_7 + \widehat{\bm{F}}^a,
\end{equation}
where $\bm{W}^a_6$ is the weight matrix of the GCN layer, $\bm{W}^a_7$ is the weight matrix of residual structure. 
The output $\widetilde{\bm{F}}^a=\{\widetilde{\bm{f}}_{t,k}^a\}_{t=1,k=1}^{t=T,k=K} \in \mathbb{R}^{(T \times K)\times D}$ is the updated features for appearance-aware objects. The updated feature $\widetilde{\bm{F}}^m$ and $\widetilde{\bm{F}}^{3d}$ for motion-aware and 3D-aware objects can be obtained in the same way.

\noindent \textbf{Object feature fusion.}
After obtaining the updated object features, we aim to integrate the object features within each frame to represent more fine-grained frame-level information under the guidance of query information. To this end, we first compute the cosine similarity between each appearance-aware object feature $\widetilde{\bm{f}}^a_{t,k}$ in frame $t$ and the sentence-level query feature $\bm{q}_{global}$ as :
\begin{equation}
    c^a_{t,k} = \frac{(\widetilde{\bm{f}}^a_{t,k})(\bm{q}_{global}\bm{W}_q)^{\top}}{||\widetilde{\bm{f}}^a_{t,k}||_2 ||\bm{q}_{global}\bm{W}_q||_2},
\end{equation}
where $\bm{W}_q$ is the projection matrix, $c^a_{t,k}$ indicates the relational score between each visual object and the given query. Then, we fuse the object features within each frame $t$ to represent current query-specific frame-level feature as:
\begin{equation}
    \bm{h}^a_t = \sum_{k=1}^K \text{softmax}(c^a_{t,k}) \widetilde{\bm{f}}^a_{t,k}.
\end{equation}
The final query-specific frame-level features in appearance, motion and 3D-aware branches can be denoted as $\bm{H}^a=\{\bm{h}^a_t\}_{t=1}^T \in \mathbb{R}^{T \times D},\bm{H}^m=\{\bm{h}^m_t\}_{t=1}^T \in \mathbb{R}^{T \times D},\bm{H}^{3d}=\{\bm{h}^{3d}_t\}_{t=1}^T \in \mathbb{R}^{T \times D}$, respectively.

\begin{figure}[t]
\centering
\includegraphics[width=0.5\textwidth]{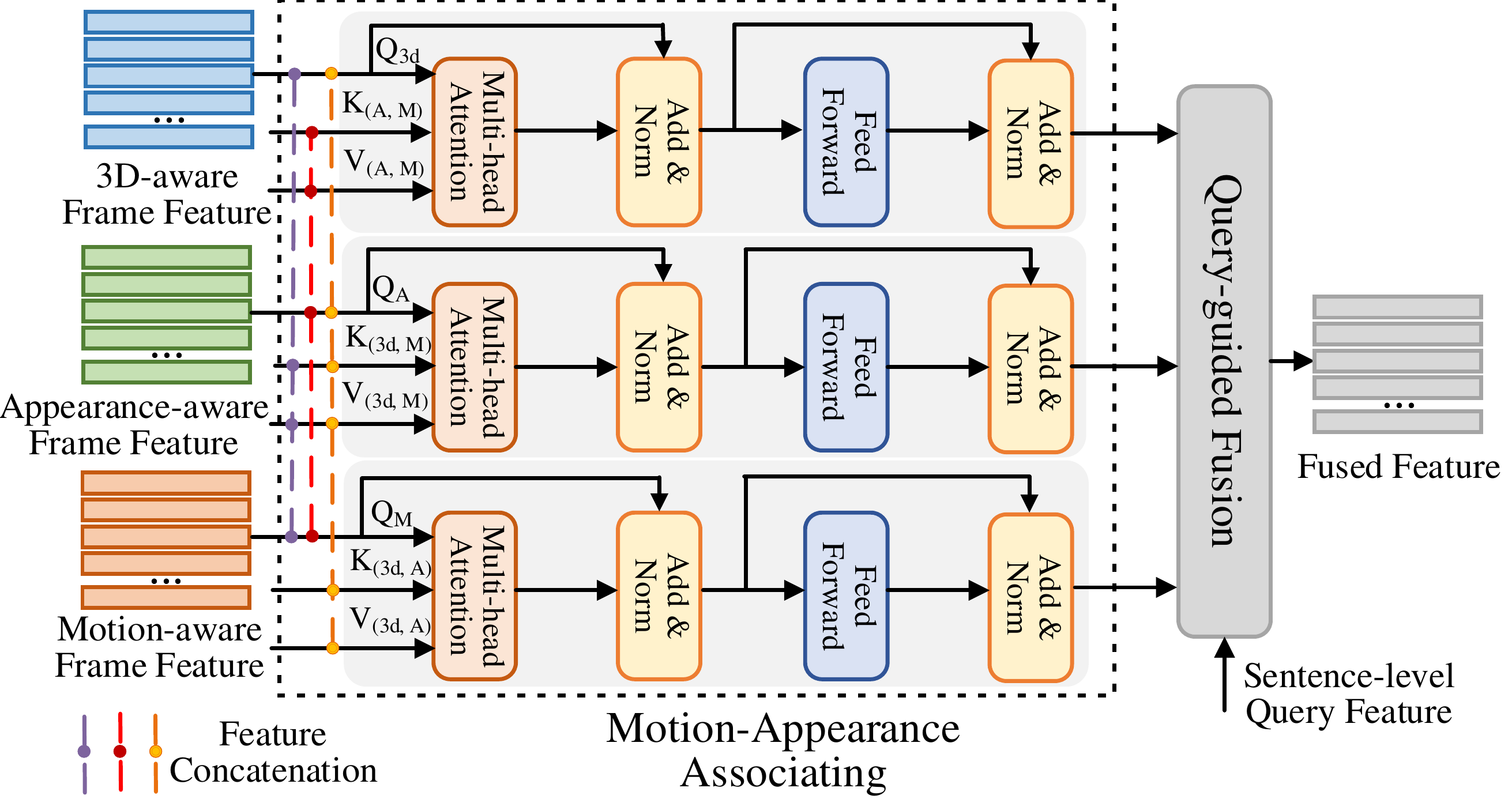}
\caption{Illustration of the motion-appearance associating module.}
\label{fig:mas}
\end{figure}

\subsection{Associating Motion and Appearance}
After generating appearance-aware, motion-aware and 3D-aware frame-level features $\bm{H}^a$, $\bm{H}^m$, and $\bm{H}^{3d}$, we develop a motion-appearance associating module to associate their features and decide which features are most discriminative for final grounding. Module details are shown in Fig.~\ref{fig:mas}.

\noindent \textbf{Appearance- and motion-guided 3D-aware feature enhancement.}
Considering the 3D-aware feature contains implicit spatial appearance and temporal motion information,  
we abstract the spatial-temporal contexts from the appearance and motion features, and integrate them into the 3D-aware features for generating 3D-aware enhanced features (details are shown in the first stream of Fig. \ref{fig:mas}). 
To incorporate the appearance and motion contexts into the 3D-aware features, we develop an attention-based mechanism to aggregate the relevant contexts for enhancing the 3D-aware features.
Inspired by transformer architecture that contains multi-head self-attention blocks for multi-inputs correlating and updating, we design a triple-modal transformer, called TriTRM, to take three kinds of features ($\bm{H}^a$, $\bm{H}^m$, and $\bm{H}^{3d}$) as input and incorporate two of them into the other one.

Generally, the transformer contains individual query ($\bm{Q}$), key ($\bm{K}$), and value ($\bm{V}$) matrices. The keys/values in our TriTRM are defined as the concatenation matrices from two modalities, the query is the matrix from the other one modality, and we fed all of them as input to the multi-head attention block to learn the triple-modal correlating and updating. As a result, the TriTRM block generates attention features to the current modality conditioned on the other two modalities. 
Typically, we define our TriTRM as follows:
\begin{equation}
\text{TriTRM}(\bm{Q},\bm{K},\bm{V})=\text{FFN}(\sigma(\frac{\bm{Q}\bm{K}^T}{\sqrt{d}}\bm{V})),
\end{equation}
where $\sigma$ denotes the softmax function, $d$ is the feature dimension of the multi-head block, and FFN denotes the feed forward network. By using such TriTRM function in our associating module, we can leverage two of the modalities as a guide to enhance the remaining one modality.

Therefore, for the associating process in the first stream in Fig. \ref{fig:mas}, we leverage the TriTRM function to obtain the enhanced 3D-aware features as follows:
\begin{equation}
\widehat{\bm{H}}^{3d}=\text{TriTRM}(\bm{Q}_{3d},\bm{K}_{(A,M)},\bm{V}_{(A,M)}),
\end{equation}
where $\widehat{\bm{H}}^{3d}$ denotes the enhanced 3D-aware features. $\bm{K}_{(A,M)},\bm{V}_{(A,M)}$ are the concatenated matrix from appearance and motion features $\bm{H}^a,\bm{H}^m$ with further linear projections.
In this soft and learnable way, contexts appeared in appearance feature and motion feature are aggregated into the 3D-aware feature space.


\noindent \textbf{3D- and motion-guided appearance enhancement.}
Compared with the 3D-aware feature $\bm{H}^{3d}$ and motion feature $\bm{H}^m$, the appearance feature $\bm{H}^a$ contains more on the appearance details and spatial location clues of a certain object. 
Similar to enhanced 3D-aware feature $\widehat{\bm{H}}^{3d}$, we also employ the TriTRM module to attend 3D-aware and motion-aware object features into appearance space for inferring appearance contexts and integrate them with the appearance features $\bm{H}^a$ as:
\begin{equation}
    \widehat{\bm{H}}^a =\text{TriTRM}(\bm{Q}_{A},\bm{K}_{(3d,M)},\bm{V}_{(3d,M)}),
\end{equation}
where $\widehat{\bm{H}}^a$ is the enhanced appearance feature. $\bm{K}_{(3d,M)},\bm{V}_{(3d,M)}$ are the concatenated contexts from 3D-aware and motion features $\bm{H}^{3d},\bm{H}^m$ with further linear projections.

\noindent \textbf{3D- and appearance-guided motion enhancement.}
Also, we can use the TriTRM module to obtain the enhanced motion feature as follows:
\begin{equation}
    \widehat{\bm{H}}^m = \text{TriTRM}(\bm{Q}_{M},\bm{K}_{(3d,A)},\bm{V}_{(3d,A)}),
\end{equation}
where $\widehat{\bm{H}}^m$ is the enhanced motion feature.

\noindent \textbf{Query-guided motion-appearance fusion.}
To determine which information is most discriminative among appearance, motion, and 3D-aware features $\widehat{\bm{H}}^a,\widehat{\bm{H}}^m,\widehat{\bm{H}}^{3d}$ corresponding to the query, we integrate them under the guidance of sentence-level query features
through the following attention-based weighted summation:
\begin{equation}
\begin{split}
    \widetilde{\bm{H}} = \ & \text{softmax}(\widehat{\bm{H}}^{3d}(\bm{q}_{global})^{\top}) \odot \widehat{\bm{H}}^{3d} \\
    &+ \text{softmax}(\widehat{\bm{H}}^a(\bm{q}_{global})^{\top}) \odot \widehat{\bm{H}}^a \\
    & + \text{softmax}(\widehat{\bm{H}}^m(\bm{q}_{global})^{\top}) \odot \widehat{\bm{H}}^m,
\end{split}
\end{equation}
where $\widetilde{\bm{H}} = \{\widetilde{\bm{h}}_t \}_{t=1}^T \in \mathbb{R}^{T \times D}$ is the integrated frame-level features to be fed into the latter grounding head.

\subsection{Grounding Head}
\begin{figure}[t]
\centering
\includegraphics[width=0.45\textwidth]{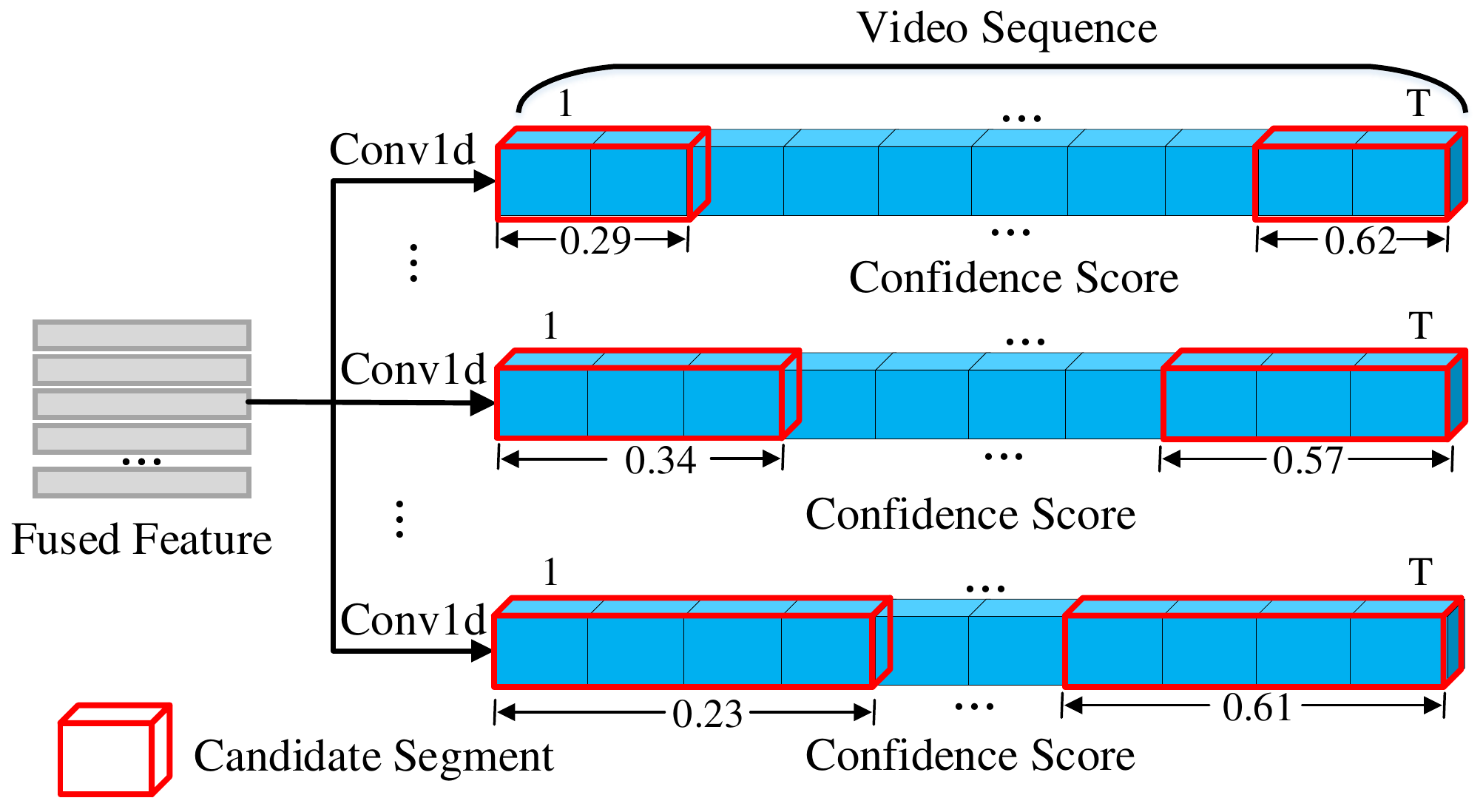}
\caption{Illustration of the proposal-ranking based grounding heads.}
\label{fig:heads}
\end{figure}

At last, we apply grounding heads on the feature $\widetilde{\bm{H}}$ to predict the target segment semantically corresponding to the query information. There are many grounding heads proposed in recent years: proposal-ranking based grounding head \cite{liu2021context,liu2020jointly,zhang2019cross,zhang2019learninga} and the boundary-regression grounding head \cite{mun2020local,chenrethinking,yuan2019find}. In this paper, we follow the former one \cite{zhang2019cross,liu2020jointly,yuan2019semantic} to determine the target video segment with pre-defined segment proposals. Specifically, we first define multiple segment proposals with different sizes on each frame $t$, and adopt multiple 1d convolutional layers to process frame-wise features $\widetilde{\bm{h}}_t$ to produce the confidence scores and temporal offsets of these segment proposals. The scoring process is shown in Fig.~\ref{fig:heads}. Supposing there are $R$ proposals within the entire video, for each segment proposal whose start and end timestamp is $(\tau_s,\tau_e)$, we denote its confidence score and boundary offsets as $o$ and $(\delta_s,\delta_e)$, where $s,e$ means the start and end. Thus, the predicted segments of each proposal can be presented as $(\tau_s+\delta_s,\tau_e+\delta_e)$. During training, we compute the Intersection over Union (IoU) score $o^{gt}$ between each pre-defined segment proposal and the ground truth, and utilize it to supervise the confidence score as:
\begin{equation}
    \mathcal{L}_{iou} = - \frac{1}{R} \sum o^{gt} \text{log}(o)+(1-o^{gt})\text{log}(1-o).
\end{equation}
As the boundaries of pre-defined segment proposals are relatively coarse, we only fine-tune the localization offsets of positive segments by a boundary loss. In details, if the confidence score $o$ of each proposal is larger than a threshold value $\lambda$, we define this proposal is a positive sample. We promote localizing precise start and end points by the boundary loss as follows:
\begin{equation}
    \mathcal{L}_{boundary} = \frac{1}{R_{pos}} \sum \mathcal{L}_1(\delta_s - \delta_s^{gt}) + \mathcal{L}_1(\delta_e - \delta_e^{gt}),
\end{equation}
where $R_{pos}$ is the number of positive samples, $\mathcal{L}_1$ denotes the smooth L1 function. The overall loss function can be formulated as:
\begin{equation}
    \mathcal{L} = \mathcal{L}_{iou} + \alpha \mathcal{L}_{boundary},
\end{equation}
where parameter $\alpha$ is utilized to control the balance. 

\section{Experiments}
\subsection{Datasets and Evaluation Metric}
We conduct experiments on three challenging benchmark datasets: ActivityNet Caption \cite{krishna2017dense}, Charades-STA \cite{gao2017tall}, and TACoS \cite{regneri2013grounding}, summarized in Table~\ref{tab:dataset}.

\begin{table*}[t!]
    \small
    \centering
    \caption{Summaries of ActivityNet Caption, Charades-STA and TACoS Datasets, including data domain, number of training/validation/testing samples, average video duration, average target moment duration and average query length.}
    \begin{tabular}{cccccccc}
    \hline \hline
    Dataset & Domain & Training Set & Validation Set & Testing Set & Video Time & Moment Time & Query Length \\ \hline
    ActivityNet Caption \cite{krishna2017dense} & Open & 37421 & 17505 & 17031 & 117.61s & 36.18s & 14.78 \\
    Charades-STA \cite{gao2017tall} & Indoor & 12408 & - & 3720 & 30.59s & 8.22s & 7.22 \\
    TACoS \cite{regneri2013grounding} & Cooking & 10146 & 4589 & 4083 & 287.14 & 5.45 & 10.05 \\ \hline
    \end{tabular}
    \label{tab:dataset}
\end{table*}

\noindent \textbf{ActivityNet Caption.}
Activity Caption \cite{krishna2017dense} contains 20000 untrimmed videos with 100000 descriptions from YouTube \cite{caba2015activitynet}. The videos are 2 minutes on average, and these annotated video clips have much larger variation, ranging from several seconds to over 3 minutes. Since the test split is withheld for competition, following public split \cite{gao2017tall}, we use 37421, 17505, and 17031 sentence-video pairs for training, validation, and testing respectively.

\noindent \textbf{Charades-STA.}
Charades-STA is a benchmark dataset for the video grounding task, which is built upon the Charades \cite{sigurdsson2016hollywood} dataset. It is collected for video action recognition and video captioning, and contains 6672 videos and involves 16128 video-query pairs.
Following previous work \cite{gao2017tall}, we utilize 12408 video-query pairs for training and 3720 pairs for testing.

\noindent \textbf{TACoS.}
TACoS is collected by \cite{regneri2013grounding} for video grounding and dense video captioning tasks. It consists of 127 videos on cooking activities with an average length of 4.79 minutes. In video grounding task, it contains 18818 video-query pairs. 
For fair comparisons, we follow the same split of the dataset as \cite{gao2017tall}, which has 10146, 4589, and 4083 video-query pairs for training, validation, and testing respectively.

\noindent \textbf{Evaluation metric.}
We adopt “R@n, IoU=m” proposed by \cite{hu2016natural} as the evaluation metric, which calculates the IoU between the top-n retrieved video segments and the ground truth. It denotes the percentage of language queries having at least one segment whose IoU with ground truth is larger than m.
In our experiments, we use $n \in \{1,5\}$ for all datasets, $m \in \{0.5,0.7\}$ for ActivityNet Caption and Charades-STA, $m \in \{0.3,0.5\}$ for TACoS.

\subsection{Implementation Details}
For appearance-aware object features, 
we sample the frames in 8-frame step to avoid information redundancy.
We utilize ResNet50 \cite{he2016deep} based Faster R-CNN \cite{ren2015faster} model pretrained on Visual Genome dataset \cite{krishna2016visual} to obtain appearance-aware object features, and extract its global feature from another ResNet50 pretrained on ImageNet. The number $K$ of extracted objects is set to 20. For motion-aware object features, we compute the optical flows among the video frames by an energy minimisation framework \cite{simonyan2014two}, and extract its global features from another pretrained ResNet50 model. We apply RoIAlign \cite{he2017mask} on them to generate object features.
For 3D-aware object features,
we define continuous 16 frames as a clip and each clip overlaps 8 frames with adjacent clips. We first extract clip-wise features from a pretrained C3D \cite{tran2015learning} or I3D \cite{carreira2017quo} model, and then apply RoIAlign on them to generate object features.
Since some videos are overlong, we uniformly downsample frame- and clip-feature sequences to $T=256$.
As for sentence encoding, we utilize Glove \cite{pennington2014glove} to embed each word to 300 dimension features. The head size of multi-head self-attention is set to 8, and the hidden dimension of Bi-GRU is set to 512. The dimension $D$ is set to 1024, and the balance hyper-parameter $\alpha$ is set to 0.005. For segment proposals in grounding head, we have 800 samples for each video on both Charades-STA and TACoS datasets and 1400 samples for each video on ActivityNet Caption dataset, and set the positive threshold as $\lambda = 0.55$.
We train the whole model for 100 epochs with batch size of 16 and early stopping strategy.
Parameter optimization is performed by Adam optimizer with leaning rate $4\times 10^{-4}$ for ActivityNet Caption and Charades-STA and $3\times 10^{-4}$ for TACoS, and linear decay of learning rate and gradient clipping of 1.0. 

\begin{table}[t!]
    \centering
    \caption{Overall performance comparison among our method with the proposal-based and proposal-free methods on the ActivityNet Caption dataset under the official train/test splits.}
    \begin{tabular}{c|c|cccc}
    \hline
    \multirow{3}*{Method} & \multicolumn{5}{c}{ActivityNet Caption} \\ \cline{2-6}
    ~ & \multirow{2}*{Feature} & R@1, & R@1, & R@5, & R@5, \\ 
    ~ & ~ & IoU=0.5 & IoU=0.7 & IoU=0.5 & IoU=0.7 \\ \hline \hline
    CTRL & C3D & 29.01 & 10.34 & 59.17 & 37.54 \\
    ACRN & C3D & 31.67 & 11.25 & 60.34 & 38.57 \\
    QSPN & C3D & 33.26 & 13.43 & 62.39 & 40.78 \\
    SCDM & C3D & 36.75 & 19.86 & 64.99 & 41.53 \\
    BPNet & C3D & 42.07 & 24.69 & - & - \\
    CMIN & C3D & 43.40 & 23.88 & 67.95 & 50.73 \\
    2DTAN & VGG & 44.51 & 26.54 & 77.13 & 61.96 \\
    DRN & C3D & 45.45 & 24.36 & 77.97 & 50.30 \\
    FIAN & C3D & 47.90 & 29.81 & 77.64 & 59.66 \\
    CBLN & C3D & 48.12 & 27.60 & 79.32 & 63.41 \\ \hline \hline
    CBP & C3D & 35.76 & 17.80 & 65.89 & 46.20 \\
    GDP & C3D & 39.27 & - & - & - \\
    LGI & C3D & 41.51 & 23.07 & - & - \\ 
    VSLNet & C3D & 43.22 & 26.16 & - & - \\
    IVG-DCL & C3D & 43.84 & 27.10 & - & - \\ \hline \hline
    \multirow{2}*{\textbf{Ours}} & C3D+Object & 51.97 & 31.39 & 84.05 & 68.11 \\
    ~ & I3D+Object & \textbf{53.72} & \textbf{32.30} & \textbf{85.45} & \textbf{69.48} \\ \hline
    \end{tabular}
    \label{tab:sota1}
\end{table}

\subsection{Comparison with State-of-the-Arts}
\noindent \textbf{Compared methods.}
We compare the proposed MA3SRN with state-of-the-art TSG methods on three datasets. These methods are grouped into three categories by the viewpoints of proposal-based, proposal-free and detection-based approach: (1) Proposal-based approach: CTRL \cite{gao2017tall}, ACRN \cite{liu2018attentive}, QSPN \cite{xu2019multilevel}, SCDM \cite{yuan2019semantic}, BPNet \cite{xiao2021boundary}, CMIN \cite{zhang2019cross}, 2DTAN \cite{zhang2019learning}, DRN \cite{zeng2020dense}, FIAN \cite{qu2020fine}, CBLN \cite{liu2021context}; 
(2) Proposal-free approach: CBP \cite{wang2019temporally}, GDP \cite{chenrethinking}, LGI \cite{mun2020local}, VSLNet \cite{zhang2020span}, IVG-DCL \cite{nan2021interventional}, ACRM \cite{tang2021frame}; 
(3) Detection-based approach: MMRG \cite{zeng2021multi}. 

\noindent \textbf{Comparison on ActivityNey Caption.}
As shown in Table~\ref{tab:sota1}, we compare our MA3SRN with the state-of-the-art proposal-based and proposal-free methods on ActivityNet Caption dataset, where we achieve a new state-of-the-art performance in terms of all metrics. Particularly, the proposed C3D+Object variant outperforms the best proposal-based method CBLN with 3.85\%, 3.79\%, 4.73\% and 4.70\% improvements on the all metrics, respectively. The C3D+Object variant also makes an even larger improvement over the best proposal-free method IVG-DCL in metrics R@1, IoU=0.5 and R@1, IoU=0.5 by 8.13\% and 4.29\%.
It verifies the benefits of utilizing detection model to capture the detailed local contexts and filter out the redundant background contents among the entire video.
When using more representative I3D features, our I3D+Object variant also beats all the other methods and brings the improvement of the C3D+Object variant by 1.75\%, 1.91\%, 1.10\%, and 1.39\% over all metrics, respectively. 

\begin{table}[t!]
    \centering
    \caption{Overall performance comparison among our method with the proposal-based and proposal-free methods on the Charades-STA dataset under the official train/test splits.}
    \begin{tabular}{c|c|cccc}
    \hline
    \multirow{3}*{Method} & \multicolumn{5}{c}{Charades-STA} \\ \cline{2-6}
    ~ & \multirow{2}*{Feature} & R@1, & R@1, & R@5, & R@5, \\ 
    ~ & ~ & IoU=0.5 & IoU=0.7 & IoU=0.5 & IoU=0.7  \\ \hline \hline
    CTRL & C3D & 23.63 & 8.89 & 58.92 & 29.57  \\
    ACRN & C3D & 20.26 & 7.64 & 71.99 & 27.79  \\
    QSPN & C3D & 35.60 & 15.80 & 79.40 & 45.50  \\
    SCDM & I3D & 54.44 & 33.43 & 74.43 & 58.08 \\
    BPNet & I3D & 50.75 & 31.64 & - & -  \\
    2DTAN & VGG & 39.81 & 23.25 & 79.33 & 51.15  \\
    DRN & I3D & 53.09 & 31.75 & 89.06 & 60.05  \\
    FIAN & I3D & 58.55 & 37.72 & 87.80 & 63.52 \\
    CBLN & I3D & 61.13 & 38.22 & 90.33 & 61.69  \\ \hline \hline
    CBP & C3D & 36.80 & 18.87 & 70.94 & 50.19  \\
    GDP & C3D & 39.47 & 18.49 & - & - \\
    VSLNet & I3D & 47.31 & 30.19 & - & - \\
    IVG-DCL & I3D & 50.24 & 32.88 & - & -  \\
    ACRM & I3D & 57.53 & 38.33 & - & - \\
    \hline \hline
    \multirow{2}*{\textbf{Ours}} & C3D+Object & 67.23 & 45.63 & 95.14 & 73.86 \\
    ~ & I3D+Object & \textbf{68.98} & \textbf{47.79} & \textbf{96.82} & \textbf{75.41}  \\ \hline
    \end{tabular}
    \label{tab:sota2}
\end{table}

\noindent \textbf{Comparison on Charades-STA.}
We also compare MA3SRN with the state-of-the-art proposal-based and proposal-free methods on the Charades-STA dataset in Table \ref{tab:sota2}, where we reach the highest results over all evaluation metrics. Particularly, our C3D+Object variant outperforms the best proposal-based method CBLN by 7.41\% and 12.17\% absolute improvement in terms of R@1, IoU=0.7 and R@5, IoU=0.7, respectively. Compared to the proposal-free method ACRM, the C3D+Object model outperforms it by 9.70\% and 7.30\% in terms of R@1, IoU=0.5 and R@1, IoU=0.7, respectively.
To make a fair comparison with the existing detection-based method MMRG, as shown in Table~\ref{tab:sota4}, we follow the same data splits for training/testing. It shows that our C3D+Object variant outperforms MMRG in all metrics by a large margin. The good performance of our model is attributed to the additional temporal modeling of the optical-flow-guided motion learning beside the detection model. We further utilize the I3D to present a new I3D+Object variant, which performs better than C3D+Object since I3D can obtain stronger features.

\noindent \textbf{Comparison on TACoS.}
Table \ref{tab:sota3} and Table \ref{tab:sota4} also report the comparison of grounding results on TACoS dataset. Compared to the proposal-based method CBLN, our C3D+Object model outperforms it by 8.90\%, 10.00\%, 6.06\%, and 8.03\% in terms of all metrics, respectively. Our model also outperforms the proposal-free method ACRM by 9.09\% and 10.71\% in terms of R@1, IoU=0.3 and R@1, IoU=0.5, respectively. Compared to the detection-based method MMRG, our C3D+Object model brings the improvements of 5.92\%, 5.13\%, 5.31\% and 5.94\% in all metrics, respectively. The I3D+Object variant further achieves better results.


\begin{table}[t!]
    \centering
    \caption{Overall performance comparison among our method with the proposal-based and proposal-free methods on the TACoS dataset under the official train/test splits.}
    \begin{tabular}{c|c|cccc}
    \hline
    \multirow{3}*{Method} & \multicolumn{5}{c}{TACoS} \\ \cline{2-6}
    ~ & \multirow{2}*{Feature} & R@1, & R@1, & R@5, & R@5, \\ 
    ~ & ~ & IoU=0.3 & IoU=0.5 & IoU=0.3 & IoU=0.5  \\ \hline \hline
    CTRL & C3D & 18.32 & 13.30 & 36.69 & 25.42 \\
    ACRN & C3D & 19.52 & 14.62 & 34.97 & 24.88 \\
    QSPN & C3D & 20.15 & 15.32 & 36.72 & 25.30 \\
    SCDM & C3D & 26.11 & 21.17 & 40.16 & 32.18 \\
    BPNet & C3D & 25.96 & 20.96 & - & - \\
    CMIN & C3D & 24.64 & 18.05 & 38.46 & 27.02 \\
    2DTAN & C3D & 37.29 & 25.32 & 57.81 & 45.03 \\
    DRN & C3D & - & 23.17 & - & 33.36 \\
    FIAN & C3D & 33.87 & 28.58 & 47.76 & 39.16 \\
    CBLN & C3D & 38.98 & 27.65 & 59.96 & 46.24 \\ \hline \hline
    CBP & C3D & 27.31 & 24.79 & 43.64 & 37.40 \\
    GDP & C3D & 24.14 & - & - & - \\
    VSLNet & C3D & 29.61 & 24.27 & - & - \\
    IVG-DCL & C3D & 38.84 & 29.07 & - & - \\
    ACRM & I3D & 38.79 & 26.94 & - & - \\
    \hline \hline
    \multirow{2}*{\textbf{Ours}} & C3D+Object & 47.88 & 37.65 & 66.02 & 54.27 \\
    ~ & I3D+Object & \textbf{49.41} & \textbf{39.11} & \textbf{67.26} & \textbf{55.90} \\ \hline
    \end{tabular}
    \label{tab:sota3}
\end{table}

\subsection{Efficiency Comparison}
\begin{figure}[t!]
\centering
\includegraphics[width=0.25\textwidth]{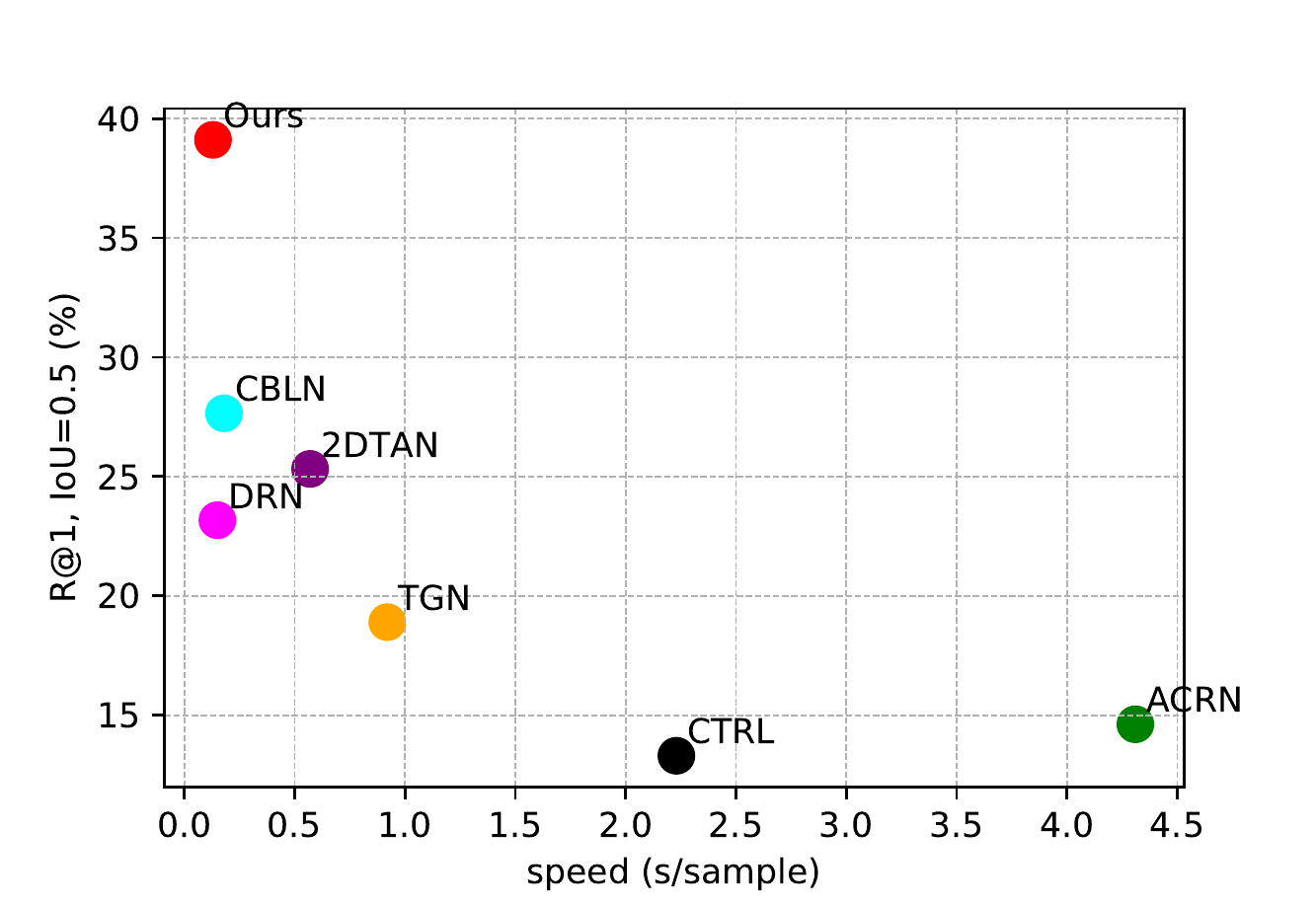}
\hspace{-0.2in}
\includegraphics[width=0.25\textwidth]{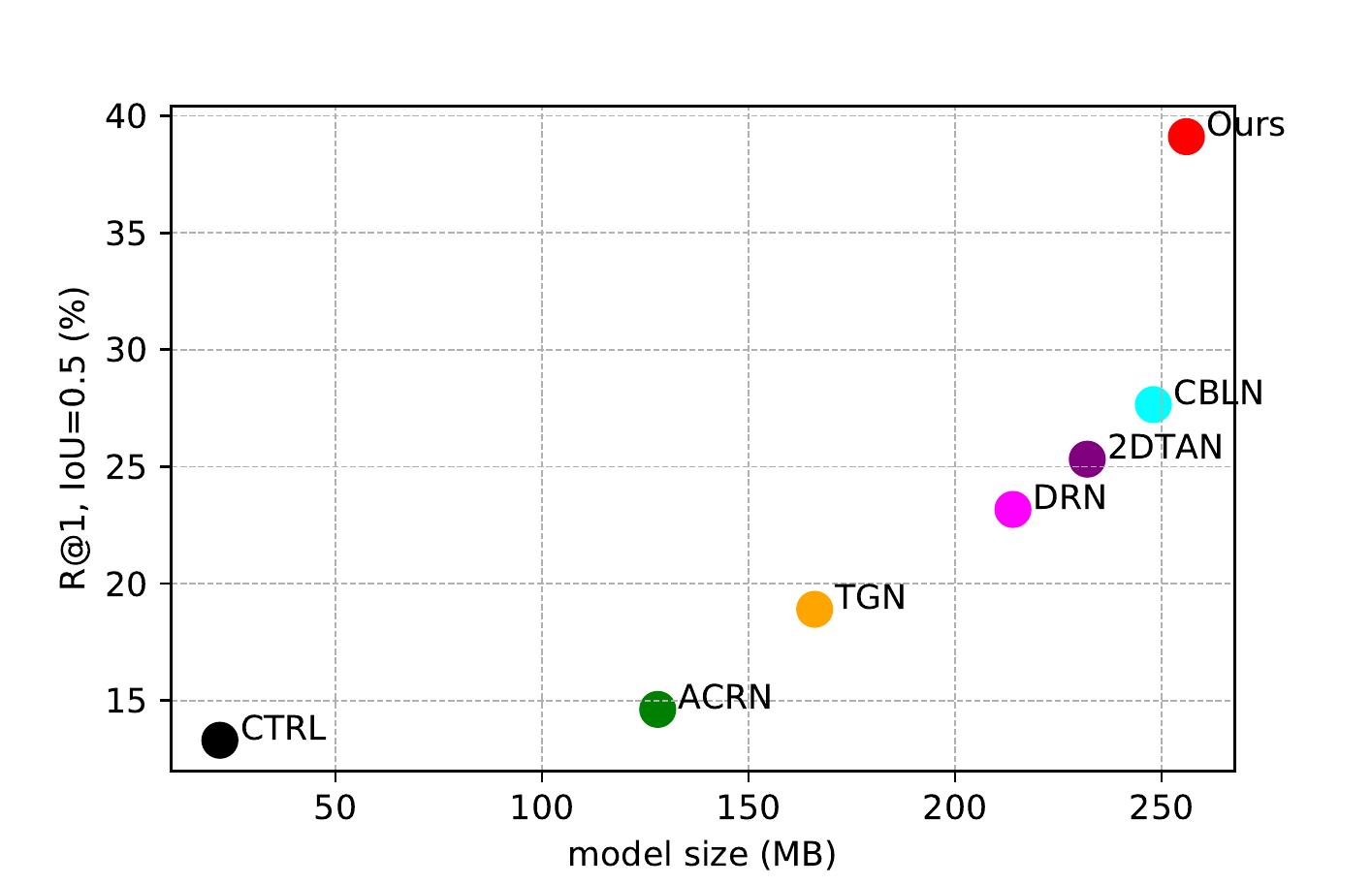}
\caption{Comparisons on speed (the average time to localize one sentence in a given video) and model size with other baselines on TACoS dataset.}
\label{fig:efficiency}
\end{figure}

We further evaluate the efficiency of our MA3SRN, by fairly comparing its running speed and parameter size with existing methods on TACoS dataset.
Particularly, since the detection process is implemented offline, we only compute inference speed of our grounding model.
As shown in Fig.~\ref{fig:efficiency}, the ``speed (s/sample)" denotes the average time to localize one sentence in a given video, ``model size (MB)" denotes the size of parameters.
It can be observed that we achieve the state-of-the-art grounding performance (R@1, IoU=0.5) with much faster processing speeds and similar parameters sizes. This attributes to: (1) Since CTRL and ACRN need to sample candidate segments with various sliding windows, they need a quite time-consuming matching procedure. (2) 2DTAN and DRN adopt much convolutional layers to achieve multi-step feature fusion, which contains a large number of parameters and are cost time. (3) CBLN utilizes multiple global and local windows to extract different level contexts among the entire video, thus needs more parameters.
(4) Our MA3SRN is free from above complex and time-consuming operations, showing superiority in both effectiveness and efficiency. Besides, our model size is similar to CBLN.


\begin{table}[t!]
    \centering
    \caption{Comparison with detection-based method MMRG on Charades-STA and TACoS datasets under MMRG's train/test splits. We do not compare with \cite{zhang2020object,zhang2020does} since they address different tasks and datasets and are close source.}
    \begin{tabular}{c|c|cccc}
    \hline \hline
    \multicolumn{6}{c}{Charades-STA} \\ \hline
    \multirow{2}*{Method} & \multirow{2}*{Feature} & R@1, & R@1, & R@5, & R@5,  \\ 
    ~ & ~ & IoU=0.5 & IoU=0.7 & IoU=0.5 & IoU=0.7 \\ \hline
    MMRG & Object & 44.25 & - & 60.22 & - \\ \hline
    \multirow{2}*{\textbf{Ours}} & C3D+Object & 49.88 & 36.09 & 67.45 & 43.27  \\
    ~ & I3D+Object & \textbf{51.04} & \textbf{37.28} & \textbf{68.97} & \textbf{45.16} \\ \hline \hline
    \multicolumn{6}{c}{TACoS} \\ \hline
    \multirow{2}*{Method} & \multirow{2}*{Feature} & R@1, & R@1, & R@5, & R@5,  \\ 
    ~ & ~ & IoU=0.5 & IoU=0.7 & IoU=0.5 & IoU=0.7 \\  \hline
    MMRG & Object & 57.83 & 39.28 & 78.38 & 56.34 \\ \hline
    \multirow{2}*{\textbf{Ours}} & C3D+Object & 63.75 & 44.41 & 83.69 & 62.28  \\
    ~ & I3D+Object & \textbf{65.42} & \textbf{46.10} & \textbf{86.18} & \textbf{63.66} \\ \hline
    \end{tabular}
    \label{tab:sota4}
\end{table}

\begin{table*}[t!]
    \centering
    \caption{Main ablation study of the proposed MA3SRN under the official train/test splits on Charades-STA dataset. It investigates three-stream feature encoders, three-stream reasoning branches, and the motion-appearance associating module.}
    \setlength{\tabcolsep}{1.6mm}{
    \begin{tabular}{c|cccccccccc|cccc}
    \hline \hline
    \multirow{2}*{Model} & & \multicolumn{3}{c}{Feature Encoding} & & \multicolumn{3}{c}{Reasoning Branches} & & Associating & R@1, & R@1, & R@5, & R@5, \\ \cline{3-5} \cline{7-9}
    ~ & & 3D-aware & Appearance & Motion & & 3D-aware & Appearance & Motion & & Module & IoU=0.5 & IoU=0.7 & IoU=0.5 & IoU=0.7 \\ \hline
    \ding{172} & & $\checkmark$ & $\times$ & $\times$ & & $\times$ & $\times$ & $\times$ &  & $\times$ & 59.79 & 36.36 & 86.68 & 64.57 \\
    \ding{173} & & $\checkmark$ & $\times$ & $\times$ & & $\checkmark$ & $\times$ & $\times$ &  & $\times$ & 61.45 & 38.96 & 89.74 & 67.29 \\
    \ding{174} & & $\times$ & $\checkmark$ & $\times$ & & $\times$ & $\times$ & $\times$ &  & $\times$ & 57.84 & 34.76 & 85.27 & 63.09  \\
    \ding{175} & & $\times$ & $\checkmark$ & $\times$ & & $\times$ & $\checkmark$ & $\times$ &  & $\times$ & 60.28 & 37.20 & 88.44 & 65.97 \\ \hline
    \ding{176} & & $\checkmark$ & $\checkmark$ & $\times$ & & $\checkmark$ & $\checkmark$ & $\times$ &  & $\times$ & 62.85 & 40.85 & 91.56 & 69.19 \\
    \ding{177} & & $\checkmark$ & $\checkmark$ & $\times$ & & $\checkmark$ & $\checkmark$ & $\times$ &  & $\checkmark$ & 64.47 & 43.09 & 93.16 & 71.55 \\
    \ding{178} & & $\checkmark$ & $\times$ & $\checkmark$ & & $\checkmark$ & $\times$ & $\checkmark$ &  & $\times$ & 62.71 & 40.47 & 90.96 & 68.62 \\
    \ding{179} & & $\checkmark$ & $\times$ & $\checkmark$ & & $\checkmark$ & $\times$ & $\checkmark$ &  & $\checkmark$ & 64.22 & 42.60 & 92.37 & 70.40\\
    \ding{180} & & $\times$ & $\checkmark$ & $\checkmark$ & & $\times$ & $\checkmark$ & $\checkmark$ &  & $\times$ & 61.92 & 38.66 & 89.59 & 67.44 \\
    \ding{181} & & $\times$ & $\checkmark$ & $\checkmark$ & & $\times$ & $\checkmark$ & $\checkmark$ &  & $\checkmark$ & 63.33 & 41.54 & 91.76 & 68.48 \\ \hline
    Full & & $\checkmark$ & $\checkmark$ & $\checkmark$ & & $\checkmark$ & $\checkmark$ & $\checkmark$ &  & $\checkmark$ & \textbf{67.23} & \textbf{45.63} & \textbf{95.14} & \textbf{73.86} \\ \hline
    \end{tabular}}
    \label{tab:ablation1}
\end{table*}

\subsection{Ablation Study}
In this section, we will perform in-depth ablation studies to examine the effectiveness of each component in our MA3SRN on Charades-STA dataset. Since most previous works utilize C3D to extract features in this task and our C3D+object variant already achieves the state-of-the-art performance, to neglect the impact of strong feature extractor I3D, we only utilize the C3D+Object variant as our backbone here.

\begin{table}[t!]
    \centering
    \caption{Ablation study on the video and query encoders on Charades-STA dataset.}
    \setlength{\tabcolsep}{1.0mm}{
    \begin{tabular}{c|c|cccc}
    \hline \hline
    \multirow{2}*{Module} & \multirow{2}*{Changes} & R@1, & R@1, & R@5, & R@5,  \\
     ~ & ~ & IoU=0.5 & IoU=0.7 & IoU=0.5 & IoU=0.7 \\ \hline
     \multirow{4}*{\tabincell{c}{Video\\Encoder}} & w/ global feature & \textbf{67.23} & \textbf{45.63} & \textbf{95.14} & \textbf{73.86} \\ 
     ~ & w/o global feature & 65.08 & 43.41 & 92.77 & 71.09 \\ \cline{2-6}
     ~ & w/ position encoding & \textbf{67.23} & \textbf{45.63} & \textbf{95.14} & \textbf{73.86} \\
     ~ & w/o position encoding & 63.68 & 42.44 & 92.30 & 70.72 \\ \hline
     \multirow{4}*{\tabincell{c}{Query\\Encoder}} & w/ self-attention & \textbf{67.23} & \textbf{45.63} & \textbf{95.14} & \textbf{73.86} \\ 
     ~ & w/o self-attention & 65.43 & 43.96 & 93.17 & 71.55 \\ \cline{2-6}
     ~ & w/ Bi-GRU & \textbf{67.23} & \textbf{45.63} & \textbf{95.14} & \textbf{73.86} \\
     ~ & w/o Bi-GRU & 65.61 & 44.20 & 93.42 & 71.92 \\ \hline
    \end{tabular}}
    \label{tab:ablation2}
\end{table}

\noindent \textbf{Main ablation studies.}
We first conduct an main ablation studies to examine the effectiveness of all the modules in our model, including different feature encoders (3D-aware, appearance, motion), different reasoning branches (3D-aware, appearance, motion), and the motion-appearance associating module.
The corresponding results of the ablation study are reported in Table~\ref{tab:ablation1}.
Model \ding{172} and model \ding{174} are the baseline models, where we only utilize a general C3D model and a ResNet50 based Faster-RCNN for 3D-aware and appearance-aware object feature extraction, respectively. We do not consider motion-aware baseline since it lacks appearance semantics for activity modelling.
Instead of building corresponding reasoning branch in model \ding{172} and \ding{174}, we directly employ a general co-attention \cite{lu2016hierarchical} module to interact object-level cross-modal information and simply concatenate query-object features for semantic enhancement. We also utilize another co-attention module to capture object-relations, and a mean-pooling layer to fuse object features to represent the frame-level features. We utilize the same grounding head in all ablation variants.
From the Table \ref{tab:ablation1}, we can summary the following conclusions: 
(1) It is worth noticing that the baseline models \ding{172} and \ding{174} achieves better performance than almost all existing methods in Table \ref{tab:sota2}, demonstrating that the detection-based method is more effective in distinguishing the frames with high similarity. That means that our object-level features are able to filter out redundant background information in frame-level features of previous works, thus leading to fine-grained activity understanding and more precise grounding. Besides, the baseline model \ding{172} performs better than the baseline model \ding{174}, as the C3D features capture both potential spatial appearance and temporal motion information and are more contextual than the detection-based appearance ones.
(2) By adding the reasoning branches to the baseline models \ding{172} and \ding{174}, models \ding{173} and \ding{175} boost a lot since the corresponding reasoning branches help to highlight the query-relevant objects while weaken the irrelevant ones. Besides, the reasoning branches also provide a graph network for more fine-grained object correlating, leading to better activity modelling.
(3) By joint learning two kinds of video features (3D-aware and appearance, 3D-aware and motion, appearance and motion) in models \ding{176}, \ding{178}, \ding{180}, the grounding performance improves a lot. This is because the motion, appearance, and 3D-aware features are complementary to each other. Once the action-oriented motion contexts are incorporated into appearance-based/3D-aware features or the appearance-aware contexts are incorporated into motion-based/3D-aware features, the model can generate more contextual and representative features for final grounding.
Besides, the motion-appearance association module further brings large improvement in models \ding{177}, \ding{179}, \ding{181}, which proves the effectiveness of incorporating appearance, motion and 3D-aware features for bi-directional enhancement.
(4) Joint learning all three kinds of video features (3D-aware, appearance and motion) achieves the best results, the performance boost is larger than using only two of them, demonstrating that the motion, appearance, and 3D-aware features are complementary to each other.

\noindent \textbf{Analysis on the multi-modal encoders.}
As shown in Table \ref{tab:ablation2}, 
We conduct the investigation on different variants of multi-modal encoders. 
For video encoding, we find that the full model performs worse (degenerate 2.15\%, 2.22\%, 2.37\% and 2.77\% in all metrics) if we remove the global feature in three branches. It demonstrates that the global feature helps to better explore the non-local object information among the objects in the same frame. 
Besides, it also presents the effectiveness of the position encoding in identifying spatial-temporal knowledge and improving the accuracy of 3.55\%, 3.19\%, 2.84\% and 3.14\%. 
For query encoding, the self-attention module is a kind of transformer encoder that captures the intra-modality contexts. As shown in Table \ref{tab:ablation2}, it brings the improvement of 1.80\%, 1.67\%, 1.97\% and 2.31\% in all metrics. Moreover, the Bi-GRU module also helps to learn the sequential information among the word sequences, and brings additional performance (1.62\%, 1.43\%, 1.72\% and 1.94\%) to the full model.

\begin{table}[t!]
    \centering
     \caption{Ablation study on the reasoning branches on the charades-STA dataset.}
     \setlength{\tabcolsep}{1.0mm}{
    \begin{tabular}{c|c|cccc}
    \hline \hline
    \multirow{2}*{Module} & \multirow{2}*{Changes} & R@1, & R@1, & R@5, & R@5, \\
     ~ & ~ & IoU=0.5 & IoU=0.7 & IoU=0.5 & IoU=0.7 \\ \hline
     \multirow{2}*{\tabincell{c}{Cross-modal\\Interaction }} & w/ attention & \textbf{67.23} & \textbf{45.63} & \textbf{95.14} & \textbf{73.86} \\
     ~ & w/ concatenation & 65.91 & 44.22 & 93.61 & 72.38  \\ \hline
     \multirow{4}*{\tabincell{c}{Graph\\Network}} & w/ graph & \textbf{67.23} & \textbf{45.63} & \textbf{95.14} & \textbf{73.86} \\ 
     ~ & w/o graph & 65.48 & 44.06 & 93.27 & 72.15 \\ \cline{2-6}
     ~ & layer=1 & \textbf{67.23} & \textbf{45.63} & \textbf{95.14} & \textbf{73.86} \\
     ~ & layer=2 & 66.98 & 45.44 & 94.90 & 73.62 \\ \hline
     \multirow{2}*{\tabincell{c}{Object-feature\\Fusion}} & w/ attention & \textbf{67.23} & \textbf{45.63} & \textbf{95.14} & \textbf{73.86} \\ 
     ~ & w/ pooling & 65.86 & 44.29 & 94.01 & 72.35 \\ \hline
    \end{tabular}}
    \label{tab:ablation3}
\end{table}

\noindent \textbf{Analysis on the reasoning branches.}
We also conduct ablation study within the reasoning branches as shown in Table \ref{tab:ablation3}. 
For object-level cross-modal interaction, our designed attention based query-object interaction mechanism outperforms the mechanism of directly concatenating query-object features by 1.32\%, 1.41\%, 1.53\% and 1.48\%. It indicates that the simple concatenation operation fails to capture the fine-grained multi-modal semantics for latter reasoning. 
For spatio-temporal graph network, replacing the graph model with simple co-attention model will reduce the performance of 1.75\%, 1.57\%, 1,87\% and 1,71\%, since the co-attention module lacks temporal modeling while the spatio-temporal graph enables better object correlation learning. 
We also investigate the impact of different graph layer.
Table \ref{tab:ablation3} shows that our model achieves the best result when the number of graph layer is set to 1. More graph layers will result in over-smoothing problem \cite{li2018deeper}, leading to relatively lower performance that the best one.
For frame-level object-feature fusion, we find that the attention based fusion mechanism performs better than directly mean-pooling all objects features within the same frame, and outperforms the latter by 1.37\%, 1.34\%, 1.13\% and 1.51\%. It demonstrates that our attention based fusion mechanism helps to filter out the redundant object information and aggregates the most representative ones for final grounding. 

\begin{table}[t!]
    \centering
    \caption{Ablation study on the motion-appearance associating module on Charades-STA dataset.}
    \setlength{\tabcolsep}{0.4mm}{
    \begin{tabular}{cc|cc|cc|cc}
    \hline \hline
    \multicolumn{2}{c|}{3D-aware} & \multicolumn{2}{c|}{Appearance} & \multicolumn{2}{c|}{Motion} & R@1, & R@5, \\
    Appearance & Motion & 3D-aware & motion & 3D-aware & Appearance & IoU=0.7 & IoU=0.7 \\ \hline
    $\times$ & $\times$ & $\times$ & $\times$ & $\times$ & $\times$ & 42.19 & 70.74 \\
    $\checkmark$ & $\times$ & $\times$ & $\times$ & $\times$ & $\times$ & 43.56 & 72.13 \\
    $\times$ & $\checkmark$ & $\times$ & $\times$ & $\times$ & $\times$ & 43.04 & 71.83 \\
    $\checkmark$ & $\checkmark$ & $\times$ & $\times$ & $\times$ & $\times$ & 43.98 & 72.55 \\
    $\checkmark$ & $\checkmark$ & $\checkmark$ & $\checkmark$ & $\times$ & $\times$ & 45.01 & 73.47 \\
    $\checkmark$ & $\checkmark$ & $\checkmark$ & $\checkmark$ & $\checkmark$ & $\checkmark$ & \textbf{45.63} & \textbf{73.86} \\ \hline
    \end{tabular}}
    \label{tab:ablation4}
\end{table}

\noindent \textbf{Analysis on the associating module.}
As shown in Table \ref{tab:ablation4}, we further implement ablation study on the proposed motion-appearance associating module.
We define the baseline as we directly concatenate the 3D-aware, appearance and motion features without associating them. The baseline model achieves 42.19\% and 70.74\% in strict metrics R@1, IoU=0.7 and R@5, IoU=0.7, respectively.
By equipping the 3D-ware features with only
appearance information, there is a 1.37\% and 1.39\% point gain compared to the baseline. 
This is because the appearance-aware objects contribute to the locate the temporal objects in 3D features for modelling its motion.
By equipping the 3D-ware features with only
motion information, there is a 0.85\% and 1.09\% point gain compared to the baseline. 
This is because the abstracted object awareness from motion features share representative object contexts to 3D-aware features for enhancing.
By equipping the 3D-ware features with both
appearance and motion information, the model achieves better performance (43.98\% and 72.55\%).
Beside, equipping the appearance features with the 3D-aware and motion information and equipping the motion features with the 3D-aware and appearance information also contribute to generating more representative features for final grounding.

\begin{figure}[t!]
\centering
\includegraphics[width=0.25\textwidth]{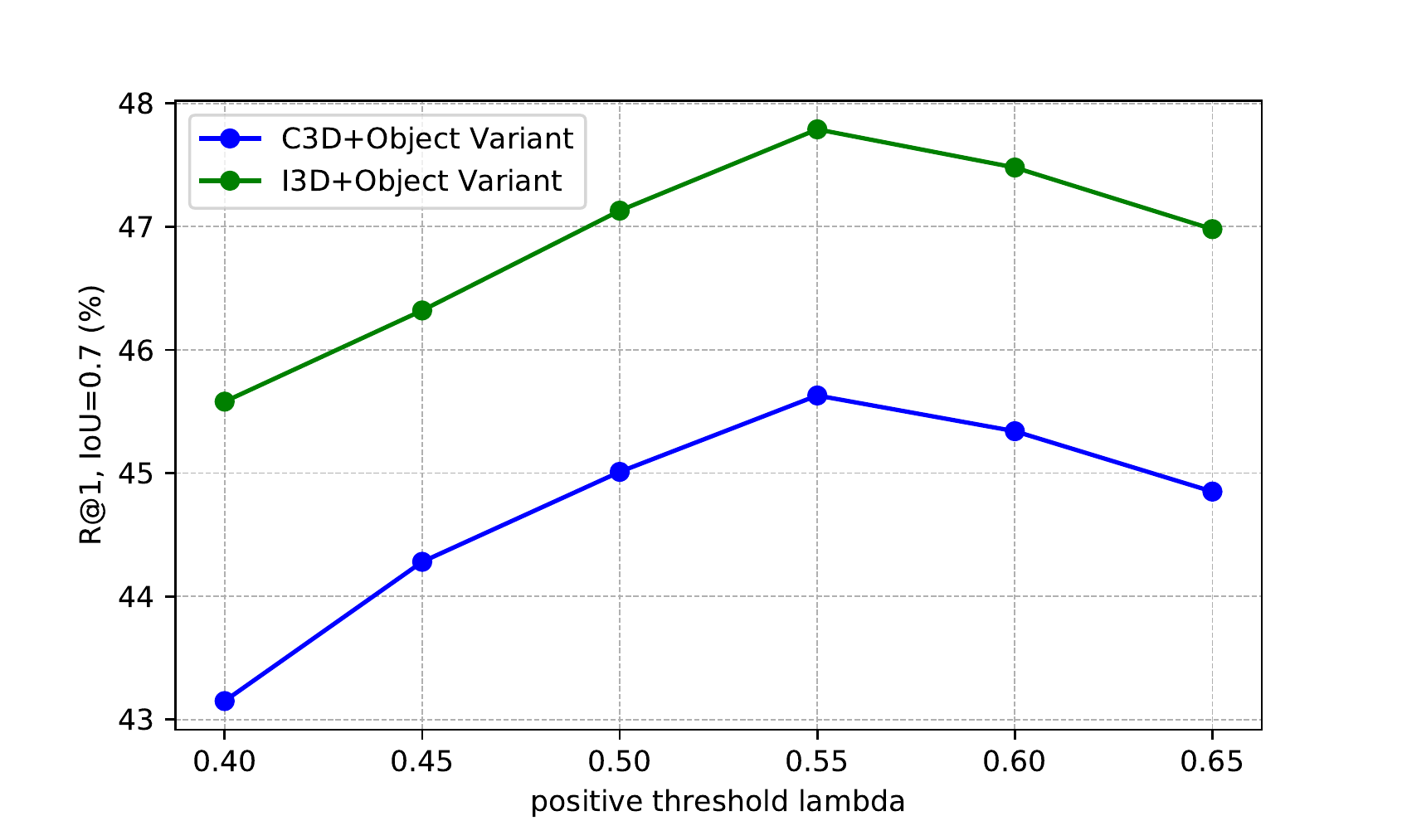}
\hspace{-0.2in}
\includegraphics[width=0.25\textwidth]{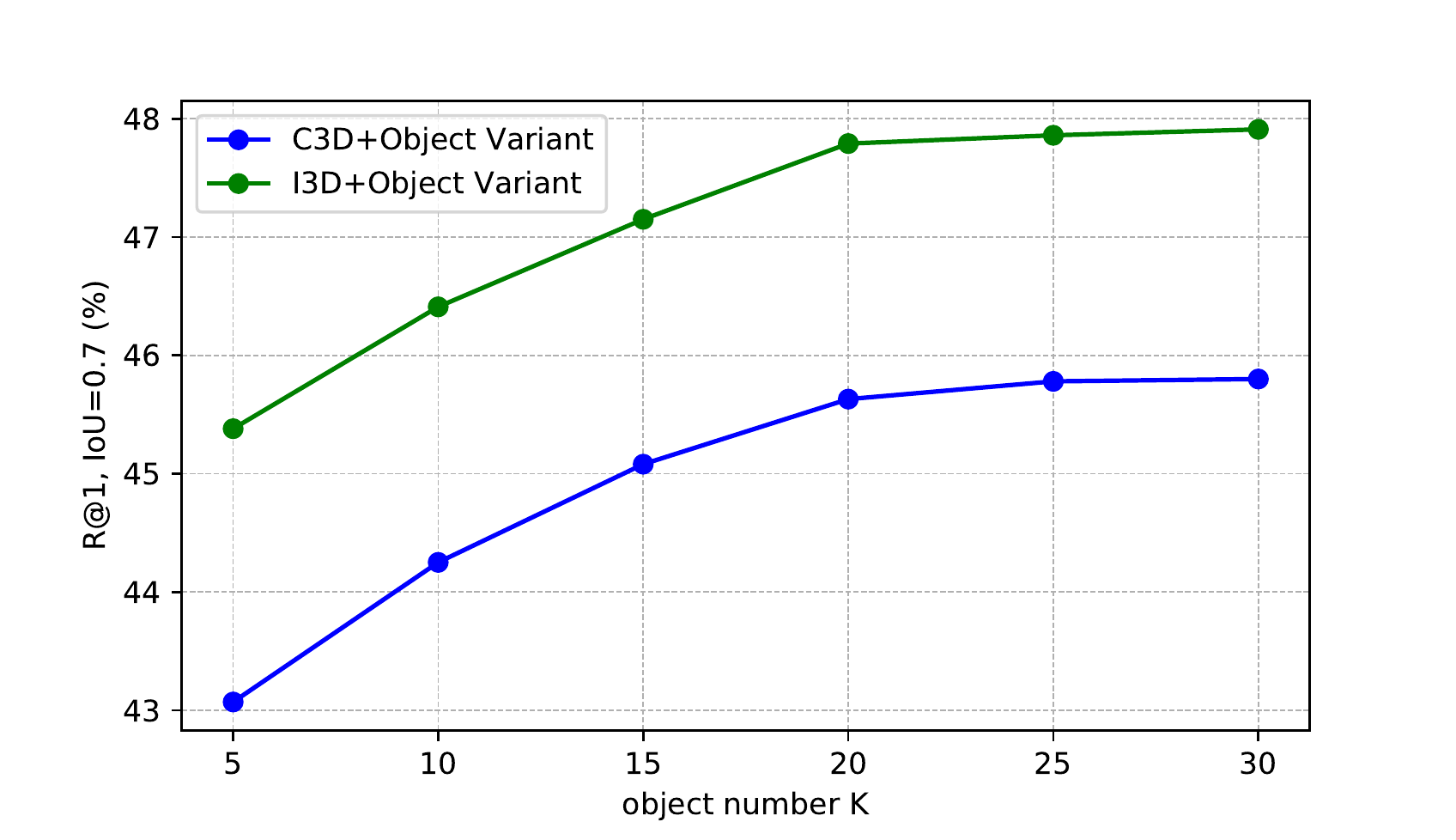}
\caption{Ablation study on the positive thredshold $\lambda$ and the object number $K$ in each frame on the Charades-STA dataset.}
\label{fig:hyperparameters}
\end{figure}

\begin{figure*}[t!]
\centering
\includegraphics[width=0.96\textwidth]{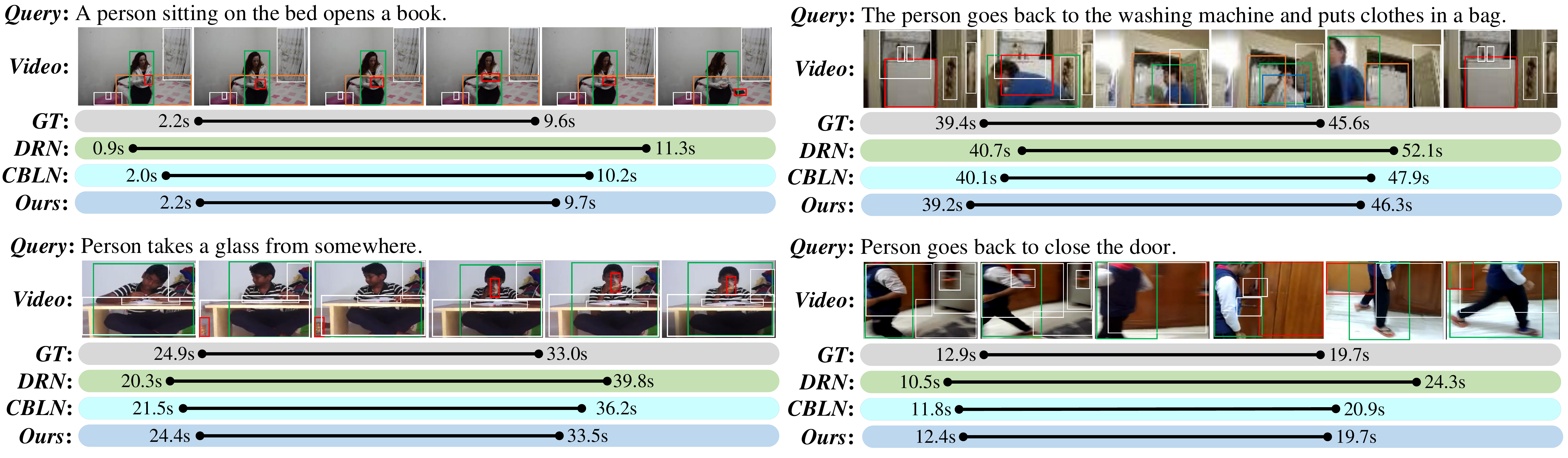}
\caption{The qualitative results of the predicted segments of different models on the Charades-STA dataset. All the grounding examples are the R@1 results.}
\label{fig:result}
\end{figure*}

\noindent \textbf{Analysis on the hyperparameters.}
As shown in Fig.~\ref{fig:hyperparameters}, we investigate the impact of two hyperparameters: positive threshold $\lambda$ and the object number $K$.
We could observe that, with the increase of $\lambda$, the variation of the performance follows a general trend,
\ie, rises at first and then starts to decline. This is because smaller $\lambda$ will guide the model take the negative segment proposals as the false positive one during grounding. In addition, larger $\lambda$ will decrease some true positive segment proposals and make the grounding much stricter, leading to worse performance.
The optimal value of $\lambda$ is 0.55, where all the variants obtain the best performance.
Besides, for the object number $K$ in each frame, the detection model will generate lots of overlapping bboxes. However, we can control both detection confidence score and non-maximum suppression threshold to reduce the number of overlapped bboxes and retain one or few bboxes for each object. Since video contents are complex, we implement different number $K$ in Fig.~\ref{fig:hyperparameters}. With larger $K$, the model performs better.
The variant with $K=30$ achieves the best result but only performs marginally better than $K=20$ at the expense of a significantly larger cost of GPU memory. Thus, we choose $K=20$ in our all experiments.

\subsection{Visualization}
To investigate the exact grounding results of our proposed methods, we provide several qualitative examples of our model and two baselines (DRN and CBLN) in Fig.~\ref{fig:result}. Here, we only show a fixed number of bounding boxes, and color the best matching ones according to the attentive weights. Since the DRN and CBLN rely on the frame-level video features that encode the whole frames, they fail to capture the subtle object details described by the sentence and filter out the complicated background visual content. Therefore, the baselines perform worse in the visualization examples. Different from them, our model learns motion-, appearance- and 3D-aware objects contexts that easily captures the appearance differences among visually similar frames, thus capture more fine-grained contexts among the video for better modelling the target activity and providing more accurate grounding results.

\section{Conclusion}
In this paper, we proposed a novel \textbf{M}otion- and \textbf{A}ppearance-guided \textbf{3}D \textbf{S}emantic \textbf{R}easoning \textbf{N}etwork (MA3SRN) for temporal sentence grounding, which incorporates optical-flow-guided motion-aware, detection-based appearance-aware, and 3D-aware object features for better reasoning spatio-temporal semantic relations between objects. Through the developed 3D-aware appearance and motion branches, our MA3SRN manages to mine 3D-aware, appearance and motion clues which match the semantic of query, and then we devise an associating module to integrate the motion-appearance information for final grounding. Experimental results on three challenging datasets (ActivityNet Caption, Charades-STA and TACoS) validate the effectiveness of our proposed MA3SRN. 

\bibliographystyle{IEEEtran}
\bibliography{reference.bib}

\end{document}